\pdfoutput=1

\documentclass[sigconf]{acmart}

\AtBeginDocument{%
  \providecommand\BibTeX{{%
    \normalfont B\kern-0.5em{\scshape i\kern-0.25em b}\kern-0.8em\TeX}}}

\setcopyright{acmcopyright}
\copyrightyear{2025}
\acmYear{2025}
\acmDOI{XXXXXXX.XXXXXXX}

\acmConference[WWW '25]{the ACM Web Conference 2025}{April 28-May 02, 2025}{Sydney, Australia}
\acmPrice{15.00}
\acmISBN{979-8-4007-XXXX-X/25/05}

\usepackage{graphicx}
\usepackage{multirow}
\usepackage{enumitem}
\usepackage{bbding}
\usepackage[capitalize]{cleveref}
\crefname{section}{Sec.}{Secs.}
\Crefname{section}{Section}{Sections}
\Crefname{table}{Table}{Tables}
\crefname{table}{Tab.}{Tabs.}

\usepackage{amsthm,amsmath}
\usepackage{mathrsfs}
\usepackage{booktabs}
\usepackage{multirow}
\usepackage{multicol}
\usepackage{color}
\usepackage{hyperref}
\usepackage{float}
\usepackage{subfig}
\usepackage{colortbl}
\usepackage{rotating}
\usepackage{array}

\definecolor{LightCyan}{rgb}{0.88,1,1}

\usepackage{xcolor}
\usepackage{adjustbox}
\usepackage{tabularx}
\usepackage{etoolbox}
\usepackage{makecell}
\usepackage[ruled]{algorithm2e}

\SetAlgoNlRelativeSize{0}

\definecolor{sgreen}{RGB}{30, 150, 30}

\begin{document}

\title{Towards Emotion Analysis in Short-form Videos: A Large-Scale Dataset and Baseline}

\author{Xuecheng Wu\textsuperscript{1}, Heli Sun\textsuperscript{1$\ddagger$}, Junxiao Xue\textsuperscript{2}, Jiayu Nie\textsuperscript{1}\authornotemark[1], Xiangyan Kong\textsuperscript{3}\authornotemark[1], Ruofan Zhai\textsuperscript{4}\authornotemark[1], Liang He\textsuperscript{1}\authornotemark[1]}

\affiliation{
\textsuperscript{1}School of Computer Science and Technology, Xi'an Jiaotong University, Xi'an, China
\\\textsuperscript{2}Research Center for Space Computing System, Zhejiang Lab, Hangzhou, China
\\\textsuperscript{3}School of Electronics and Information Engineering, Harbin Institute of Technology, Harbin, China
\\\textsuperscript{4}School of Cyber Science and Engineering, Zhengzhou University, Zhengzhou, China
\country{}}
\email{
wuxc3@stu.xjtu.edu.cn, hlsun@xjtu.edu.cn}

\renewcommand{\shortauthors}{Xuecheng Wu et al.}

\begin{abstract}
Nowadays, short-form videos (SVs) are essential to web information acquisition and sharing in our daily life. The prevailing use of SVs to spread emotions leads to the necessity of conducting video emotion analysis (VEA) towards SVs. Considering the lack of SVs emotion data, we introduce a large-scale dataset named eMotions, comprising 27,996 videos. Meanwhile, we alleviate the impact of subjectivities on labeling quality by emphasizing better personnel allocations and multi-stage annotations. In addition, we provide the category-balanced and test-oriented variants through targeted data sampling. Some commonly used videos, such as facial expressions, have been well studied. However, it is still challenging to analysis the emotions in SVs. Since the broader content diversity brings more distinct semantic gaps and difficulties in learning emotion-related features, and there exists local biases and collective information gaps caused by the emotion inconsistence under the prevalently audio-visual co-expressions. To tackle these challenges, we present an end-to-end audio-visual baseline AV-CANet which employs the video transformer to better learn semantically relevant representations. We further design the Local-Global Fusion Module to progressively capture the correlations of audio-visual features. The EP-CE Loss is then introduced to guide model optimization. Extensive experimental results on seven datasets demonstrate the effectiveness of AV-CANet, while providing broad insights for future works. Besides, we investigate the key components of AV-CANet by ablation studies. Datasets and code will be fully open soon. 
\end{abstract}



\begin{teaserfigure}
\vspace{-0.23cm}
\setlength{\belowcaptionskip}{-0.1cm}
\includegraphics[height=6cm, width=\textwidth]{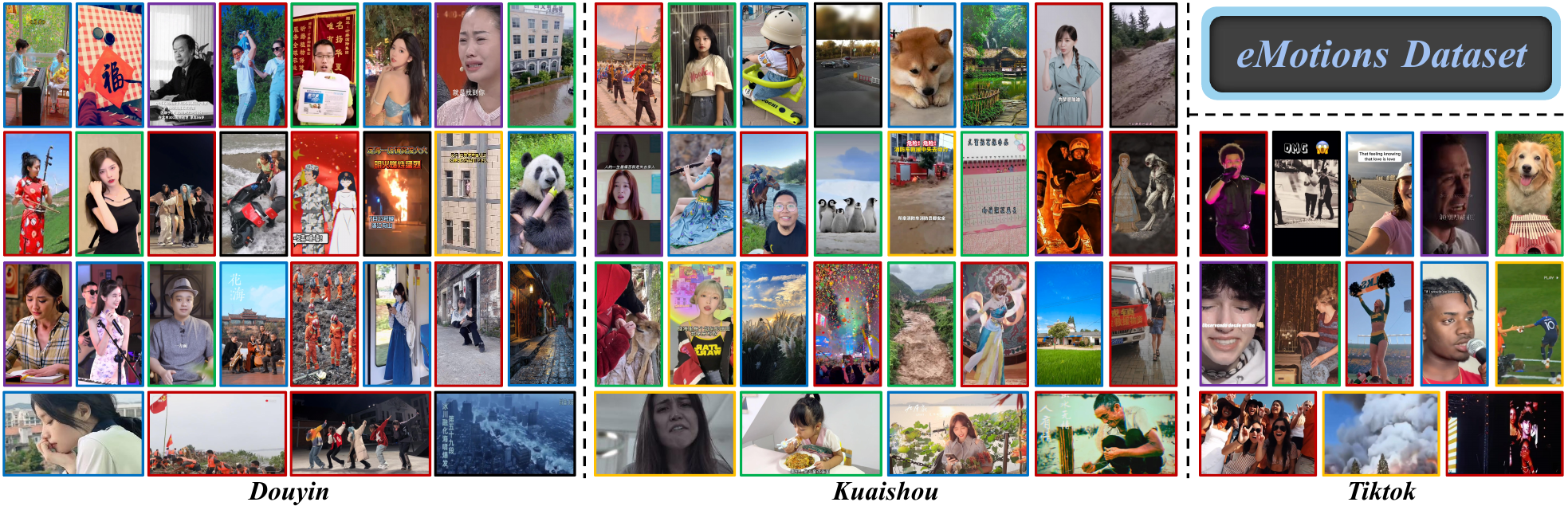}
\vspace{-2.4em}
\captionof{figure}{An overview of eMotions composed of 27,996 videos of six emotions across Douyin, Kuaishou, and Tiktok. The colors of frame borders specify the emotional categories to which they belong: \textit{\textcolor[RGB]{192,0,0}{Excitation}}, \textit{\textcolor[RGB]{0,0,0}{Fear}}, \textit{\textcolor[RGB]{0,176,80}{Neutral}}, \textit{\textcolor[RGB]{0,112,192}{Relaxation}}, \textit{\textcolor[RGB]{112,48,160}{Sadness}}, \textit{\textcolor[RGB]{255,192,0}{Tension}}.}
\label{fig1}
\end{teaserfigure}

\maketitle

\section{Introduction}
\label{sec:introduction}

Video emotion analysis (VEA) aims to mine the meanings of elements and uncover which emotion the elements evoke to the viewers. Meanwhile, the emotions of viewers can be influenced by various elements, such as videos, audio, and text from web media \cite{new_21_liu2022ser30k}, facilitating the developments of multi-modal VEA. Particularly, short-form videos\footnotemark[1] (SVs), one of the new types of social-media tools, have made rapid progress in recent years. SVs are concise and clear, combining visual, auditory, and other elements to intensify emotional expressions and arouse emotional resonance among viewers, which are crucial to spreading emotions. As a result, conducting VEA towards SVs has remarkable application values in research fields such as opinion mining \cite{new_54_sobkowicz2012opinion} and dialogue systems \cite{5_schuller2018age}. 

\footnotetext[1]{\fontsize{6.48pt}{1pt}\url{https://en.wikipedia.org/wiki/Video_clip\#Short-form_videos}}

Although VEA in facial expressions \cite{new_1_wang2020suppressing,new_3_yang2021circular} have been well studied, the research in SVs remains light because of the lack of large-scale dataset. To tackle this issue and facilitate further studies, we propose a dataset specifically constructed for emotion analysis in SVs, termed eMotions (Fig.~\ref{fig1}), which is the first large-scale dataset in this field. eMotions consists of 27,996 videos with corresponding audio from Douyin, Kuaishou, and Tiktok three SVs platforms, covering various contents across diverse dimensions and totaling almost 198 hours of durations. Specifically, considering the discrete emotional categories in psychology and the content distribution characteristics of eMotions, we label each sample using the six emotional categories proposed by Plutchik in \cite{1_plutchik1994psychology} (\textit{i.e.}, Excitation, Fear, Neutral, Relaxation, Sadness, Tension). Moreover, to alleviate the impact of subjectivities on labeling quality, we elaborately adjust the personnel allocations through proposed two-stage Cross-Check and consistencies evaluations, as well as introduce a multi-stage annotation workflow. In addition, catering to the class distribution of eMotions and the testing demands, we provide the category-balanced and test-oriented variant datasets.

\begin{figure}[]
\setlength{\belowcaptionskip}{-0.5cm}
\centering
\includegraphics[height=4.3cm, width=\linewidth]{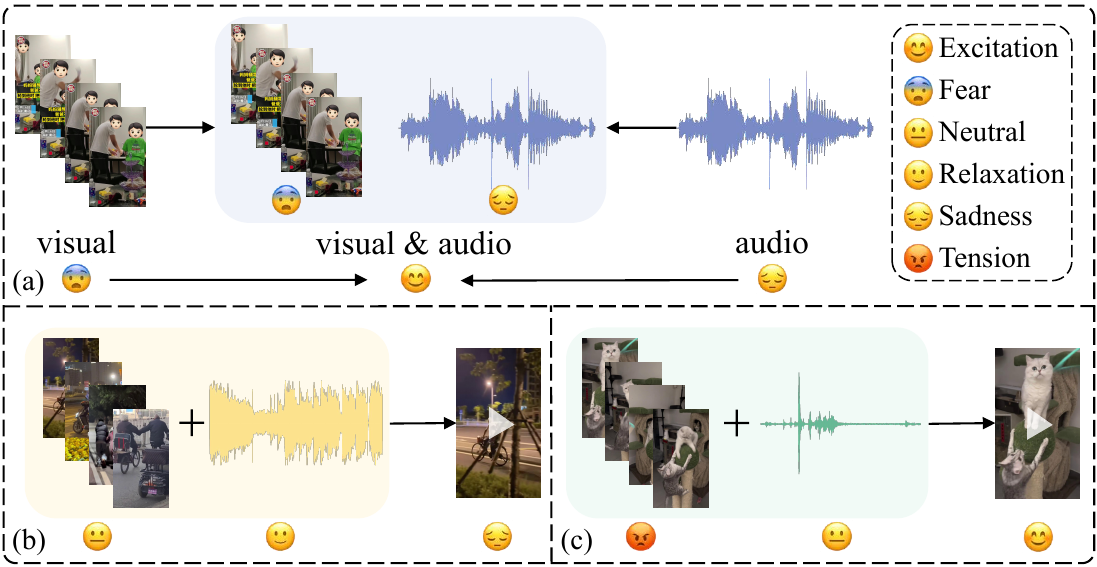}
\vspace{-1.9em}
\caption{The overall illustration of emotion inconsistence. (a) Separate visual or auditory modality evokes different emotion, leading to expression conflict. (b)\&(c) The lack of emotional information evoked from visual or auditory modality results in emotion disalignment.}
\label{fig_introd_challenge}
\end{figure}

VEA towards SVs features the following challenges: As displayed in Fig.~\ref{fig1}, the content diversity in SVs is broader, which leads to more distinct semantic gaps and barriers of learning emotion-related features than commonly used videos. Meanwhile, there exists emotion inconsistence under the prevalent audio-visual co-expressions, as shown in Fig.~\ref{fig_introd_challenge}. Therefore, at local-level, inconsistence of emotion evocations with difficulties in integrating information lead to biases in model learning. Additionally, at global-level, the obstacles in emotion understanding within the overall feature space and accumulation of local biases result in more collectively significant information gaps. Considering these observations, we analyze three keypoints as follows: (1) Learning more semantically relevant representations and emotion-related features in an end-to-end manner. (2) Interactive correlated modeling of inter-modalities at local-level, followed by selective integrations. (3) Supplementally facilitating complementary modeling of audio-visual correlations at global-level, while learning more contextual representations.

Based on these keypoints, we present an end-to-end audio-visual method denoted AV-CANet (\textbf{A}udio-\textbf{V}iusal \textbf{C}ooperatively enhanced \textbf{A}nalysis \textbf{Net}work) as the baseline on proposed eMotions dataset. Unlike previous CNNs-based VEA methods \cite{35_zhao2020end,new_4_zhou2019exploring}, we employ Video Swin-Transformer \cite{19_liu2022video} as the visual backbone, which makes AV-CANet naturally capture the global relations between regions in each frame and efficiently model the long-range dependencies as well as long-term sequences, leading to more semantically relevant representations. The Local-Global Fusion Module (LGF Module) is designed to mitigate the local biases and collective information gaps, which progressively captures the correlated information of inter-modalities to output more comprehensive representations, as displayed in Fig.~\ref{fig5-AV-CANet}. Besides, we propose the EP-CE Loss (\textbf{E}motion \textbf{P}olarity enhanced \textbf{C}ross-\textbf{E}ntropy Loss), which incorporates three emotion polarities (\textit{i.e.}, positive, neutral, negative) to guide model focusing more emotion-related features. The proposed method achieves superior performance in comparisons with advanced VEA baselines across three eMotions-related and four public datasets, indicating that it has the capability to benchmark eMotions.

In addition to VEA, the abundant emotions conveyed in eMotions can facilitate the research in LLMs (\textit{e.g.}, multi-modal alignments with the emotional behaviors of humans \cite{new_39_frisch2024llm}), emotional content generation \cite{new_55_yang2024emogen}, and explainable emotional reasoning \cite{new_56_lian2023explainable}. Our main contributions are three-fold: (1) To our knowledge, eMotions is the first large-scale dataset for VEA towards SVs. The more reliable annotated emotions can promote the developments of affective content analysis. (2) We propose the baseline AV-CANet to analysis emotions in SVs. We design corresponding components to alleviate the impact of emotion inconsistence from local to global, and leverage the emotion-polarity information to better guide model optimization. (3) We conduct extensive experiments to verify the superiority of AV-CANet and provide detailed insights into different modalities and VEA datasets for future research. Ablations are also performed to investigate the key factors of our method.

\section{Related Works}
\label{sec:related}
\subsection{Video Emotion Analysis Datasets} 
Numerous datasets have emerged to facilitate the developments of VEA. FABO \cite{new_9_gunes2006bimodal} consists of 1.9\textit{k} videos of facial and body expressions recorded by cameras. IEMOCAP \cite{26_busso2008iemocap}, the earliest audio-visual VEA dataset, originates from lab shooting. AFEW \cite{new_6_2012Collecting} contains 1,426 videos of 330 subjects labeled with seven emotions. Aff-Wild2 \cite{new_8_kollias2018aff} includes 558 videos with annotations for valence-arousal estimations. VideoEmotion8 \cite{12_jiang2014predicting} and Ekman6 \cite{13_xu2016video} consist of 1,101 and 1,637 videos drawn from video sites. CMU-MOSEI \cite{16_zadeh2018multimodal} comprises 23,500 YouTube-sourced utterances categorized under six emotions. CAER \cite{18_lee2019context} and MELD \cite{33_soujanya2018multimodal} include 13,201 and 13,708 TV shows-derived clips, respectively. Music\_video \cite{7_pandeya2021deep} houses 3,323 music recording videos labeled across six emotions. DFEW \cite{new_58_jiang2020dfew} consists of 16,372 clips from thousands of movies under seven emotions. MAFW \cite{new_57_liu2022mafw} is a compound affective facial database, including 10,045 video clips. MER2024 \cite{new_42_lian2024mer} encompasses 5,030 labeled and 115\textit{k} unlabeled videos across six emotions, setting up three subsets. Compared with these datasets, our eMotions presents the following features: (1) The first large-scale dataset for VEA towards SVs, and the currently largest audio-visual VEA dataset with full video-level annotations. (2) An emphasis on better personnel allocations and multi-stage annotations to reduce the influence of subjectivities on labeling quality, engaging 12 annotators and one expert. (3) The larger extent of content diversity, covering a multi-cultural spectrum and spanning an extensive timeline of 45 months. (4) We additionally provide two variants of eMotions for diverse needs.

\begin{figure*}[t]
\setlength{\belowcaptionskip}{-1.5em}
\centering
\includegraphics[height=7.5cm,width=\textwidth]{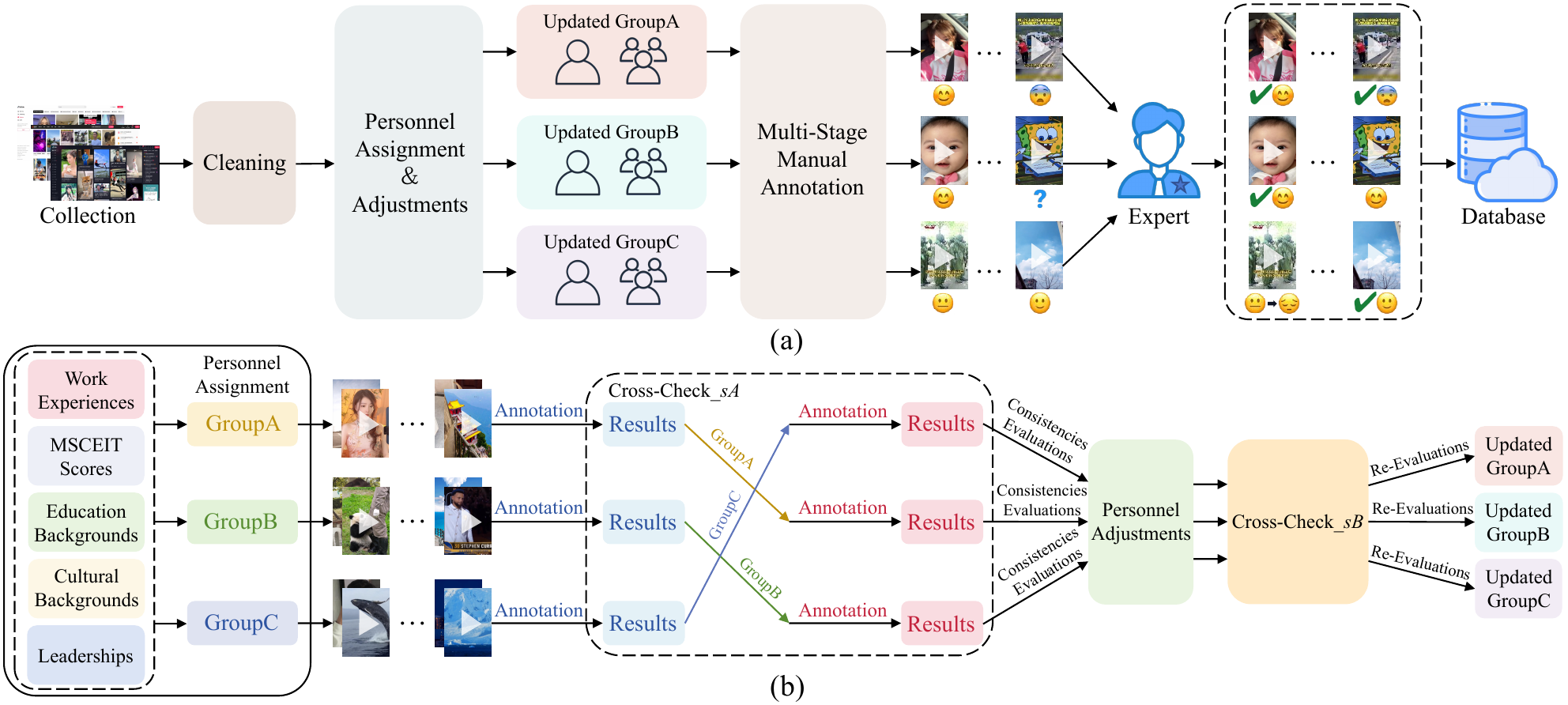}
\vspace{-2.1em}
\caption{(a) The overall pipeline of dataset construction. (b) The detailed workflow of personnel assignment and adjustments.} 
\label{fig_pipeline_adjustment}
\end{figure*}

\subsection{Video Emotion Analysis Methods} 
Early studies on VEA mainly center on designing hand-crafted features \cite{12_jiang2014predicting,new_51_hanjalic2005affective}. With the emergence of various VEA datasets, methods based on deep learning have made rapid progress. For the uni-modal methods, \cite{new_34_zhao2021former} proposes a framework consisting of the CS-Former and T-Former branches. \cite{new_12_xue2022coarse} presents a coarse-to-fine cascaded network with smooth predictions. Zhao et al. \cite{new_36_zhao2023prompting} introduces CLIP \cite{new_43_radford2021learning} and learnable tokens to enhance model performance. Regarding the audio-visual VEA approaches, \cite{40_zhang2023weakly} proposes a temporal erasing network for keyframe and context perceptions. \cite{35_zhao2020end} implements multiple attention mechanisms to output improved audio-visual features. \cite{new_13_mocanu2023multimodal} designs an intra-modality attention module to shape more refined features. \cite{new_35_su2020msaf} proposes a lightweight cross-modal fusion module that emphasizes more contributive features. \cite{new_16_tran2022pre} provides an audio-visual pre-trained framework to model the interactions between human facial and auditory behaviors. \cite{new_14_praveen2023audio} introduces a joint model to extract salient features across audio-visual modalities. \cite{50_peng2022balanced} and \cite{53_xu2023mmcosine} introduce on-the-fly gradient modulation and a novel cosine loss for performance gains, respectively. \cite{new_18_chumachenko2022self} proposes a new model for incomplete audio-visual data. \cite{55_poria2017context} presents an LSTM-based model to capture context-dependent semantics. \cite{new_59_zhang2024mart} introduce a MAE-style VEA method via masking. \cite{new_17_schoneveld2021leveraging} integrates the visual features extracted by the teacher-student networks with audio features using model-level fusion. In this work, we present AV-CANet to targetedly tackle the inherent challenges posed by the eMotions dataset, delivering superior performance across three eMotions-related and four public VEA datasets.

\section{Dataset Construction}
\label{sec:emotions}
The overall construction pipeline of eMotions is shown as Fig.~\ref{fig_pipeline_adjustment}~(a), involving data collection and cleaning (Sec.~\ref{subsec:3.1}), personnel assignment and adjustments (Sec.~\ref{subsec:3.2}), multi-stage manual annotation and expert re-review (Sec.~\ref{subsec:3.3}). We then present the labeling quality evaluations and dataset characteristics (Sec.~\ref{subsec:3.4}~\&~\ref{subsec:3.5}). Besides, we provide the details of data selections for two variants (Sec.~\ref{subsec:3.6}).

\subsection{Data Collection and Cleaning}
\label{subsec:3.1}
We collect hot events, which are diverse and representative, as the raw data. Meanwhile, we carry out a stage-by-stage crawling strategy interleaved with the formal annotation process to ensure a broad timeline. Besides, we extensively comply with the ethical considerations in terms of collection, privacy and data protections. During data cleaning, identical and corrupted videos are first removed. The videos containing racial discrimination, violence, and pornography are then eliminated through machine detection and manual review to reduce the ethical biases of trained models. Furthermore, considering the specificities of discrete emotions, videos featuring consecutive emotional shifts are also discarded. More details are provided in the appendix.

\subsection{Personnel Assignment and Adjustments}
\label{subsec:3.2}
Each annotator is asked first to pass the labeling test, comprising the sentiment quotient test and the annotation quality evaluation \cite{new_22_you2016building,new_21_liu2022ser30k}. When scoring 90 or above in accuracy, the annotators can join in the emotion labeling. We then hold training sessions for them, including the detailed introduction of annotation workflow and targeted learning on multi-cultures.

\textbf{Assignment:} As formulated in Eq.~\ref{eq1}, we empirically consider five key factors and set their coefficients to determine the assignment of group members and leaders. We also balance the gender distribution by allocating two males and one female for each group.
\vspace{-1em}
\begin{equation}
p = \alpha_\textit{1} \cdot we + \alpha_\textit{2} \cdot ms + \alpha_\textit{3} \cdot (eb + cb + lp),
\label{eq1}
\vspace{-0.05em}
\end{equation}
where $we$, $eb$, $cb$, and $lp$ respectively refer to the work experiences, education backgrounds, cultural backgrounds, and leaderships, which are all quantified using scoring tests. For MSCEIT \cite{2_mayer2002mayer} scores ($ms$), higher ranking indicates better emotional cognition. $\alpha_{i}$ ($i \in \{1, 2, 3 \}$) = $\{0.4, 0.3, 0.1\}$ are weight coefficients, respectively.

\textbf{Adjustments:} We perform personnel adjustments following assignment to alleviate the impact of subjectivities on the annotation workflow employing multi-groups with multi-annotators, which are reflected in the improvements of the consistencies of intra-group and inter-group (\textit{i.e.}, $S_a$, $S_r$) \cite{new_29_lavitas2021annotation}. Specifically, we select 9,000 samples from the cleaned data and distribute them evenly among the assigned groups. GroupA (\textit{GA}), GroupB (\textit{GB}), and GroupC (\textit{GC}) are each requested to perform annotation following the workflow described in Sec.~\ref{subsec:3.3}, and the leaders here only carry out result collections. We then sample 18 sets from the annotations of each group to conduct two-stage Cross-Check \cite{new_28_abercrombie2023consistency,new_27_hallgren2012computing}, in which we first exchange the annotations of three groups in pairs and then label these samples again, as illustrated in Fig.~\ref{fig_pipeline_adjustment}~(b). These sets, each including 100 samples, are equally divided into two parts for two-stage Cross-Check (\textit{i.e.}, \textit{sA}, \textit{sB}). Each stage has three sets for Neutral, two sets for Excitation, and one set for each of Sadness, Relaxation, Tension, as well as Fear. Afterwards, we evaluate the consistencies of intra-group and inter-group under the present allocations based on the results of Cross-Check\textit{\_sA}. The ranges of $S_a$ and $S_r$ formulated as Eq.~\ref{eq2} and Eq.~\ref{eq4} below are from 0 to 1 and 0 to 70, respectively. The larger $S_a$ and $S_r$ indicate the increasing consistencies.

\vspace{-1.15em}
\begin{align}  
S_a &= \frac{1}{n} \cdot \sum_{i=0}^{n-1} \left(\frac{1}{3 \cdot m} \cdot \sum_{j=1}^{m}c_j\right),
\label{eq2} \\[-0.37em]
S_r &= \frac{1}{c} \sum_{i=0}^{n-1} w_i \cdot\left(0.7 \cdot C_i+0.3 \cdot\left(m-{C_i}-{M_i}\right)\right), 
\label{eq4}
\end{align}
where $n$, $m$, and $c$ denote the number of sets, samples per set, and emotional categories, respectively. $c_j$ refers to the quantity of currently annotated categories of three annotators for one sample that are consistent with the previous category. $w_i \in \{\frac{1}{3} \mid 0 \leq i<3, \frac{1}{2} \mid 3 \leq i <5, 1 \mid 5 \leq i \leq 8\}$ represents the weight coefficient for each set, depending on the number of sets in each category. $C_i$ stands for the samples consistent with previous annotations. $M_i$ denotes the "more" samples, indicating the final label is indeterminate.

As shown in Tab.~\ref{tab5}, $S_a$ of the three groups achieve 0.52, 0.53, and 0.55, respectively. $S_r$ stand at 52.55, 48.78, and 54.54, with \textit{GB} scoring relatively low. Considering these observations, we perform targeted adjustments, then re-evaluate the consistencies after Cross-Check\textit{\_sB}. Following adjustments, $S_a$ of \textit{GA} and \textit{GB} both rise by 0.03, and that of \textit{GC} improves by 0.01. $S_r$ of three groups increase by 2.98, 5.63, and 1.25, in which \textit{GB} exhibits the highest improvement. These results demonstrate the effectiveness of our adjustments, indicating that we have finalized the personnel allocations that can be deployed for formal annotation. Note that the overall results of two-stage Cross-Check and the pseudocode of personnel adjustments strategy are detailed in the appendix.

\begin{table}[t]
\centering
\setlength{\abovecaptionskip}{-0em}
\renewcommand{\arraystretch}{0.7}
\setlength{\arrayrulewidth}{0.22pt}
\caption{The evaluation results of consistencies of intra-group and inter-group ($S_a$ and $S_r$) following Cross-Check.} 
\label{tab5}
\resizebox{0.4878\textwidth}{!}{%
{\tiny
\begin{tabular}{@{}ccccccc@{}}
\toprule[0.22pt]
\hspace{0.08cm}\multirow{2}{*}{\raisebox{-1ex}{Stage}} & \multicolumn{2}{c}{GroupA} & \multicolumn{2}{c}{GroupB} & \multicolumn{2}{c}{GroupC} \\ \cmidrule[0.22pt](l){2-7} 
                                & $S_a$  & $S_r$ & $S_a$  & $S_r$ & $S_a$  & $S_r$ \\ \midrule[0.22pt]
\textit{sA}                       & 0.52        & 52.55       & 0.53        & 48.78       & 0.55        & 54.54       \\ 
\textit{sB}                        & 0.55        & 55.53       & 0.56        & 54.41       & 0.56        & 55.79       \\ \bottomrule[0.22pt]
\end{tabular}
}
}
\vspace{-1.15em}
\end{table}

\subsection{Multi-Stage Manual Annotation}
\label{subsec:3.3}
The setting of categories is essential for VEA datasets. We select the six emotions proposed by Plutchik in \cite{1_plutchik1994psychology}, since they have clearer boundaries and can better capture emotional changes evidenced by Russell's Circumplex Model \cite{new_60_posner2005circumplex}. The multi-stage manual annotation workflow combines member votes and leader evaluations, benefiting to alleviate the impact of subjectivities. Following \cite{7_pandeya2021deep}, we adopt and extend the mapping table of emotion category-to-adjective to promote annotations, as shown in the appendix. Besides, we develop a labeling interface to boost the stimulation and engagement of annotators. Specifically, the leader distributes data, followed by three members undertaking annotations via the proposed mapping table. Meanwhile, members are asked to attach confidence scores to their annotations, and the average confidence score finally stands at 0.7. Next, the leader collects these annotations, leveraging a majority voting scheme to determine labels. If annotations from three members are all different (\textit{i.e.}, samples labeled "more"), the leader will intervene in labeling. If a decisive majority of four votes emerges, the final labels can be directly determined. If consensus is still unreachable, leaders will exchange samples to facilitate decision-making. In five votes, a clear majority allows us to determine the final labels. If consensus continues to be inaccessible, the expert from \href{https://cloud.baidu.com/product/dcs.html}{BD Cloud} will finalize the labels in re-review. After completing the annotation process, we calculate the overall Fleiss'kappa score, achieving $k >$ 0.45.

\begin{figure}[]
\setlength{\belowcaptionskip}{-0.5cm}
\centering
\includegraphics[height=3cm, width=\linewidth]{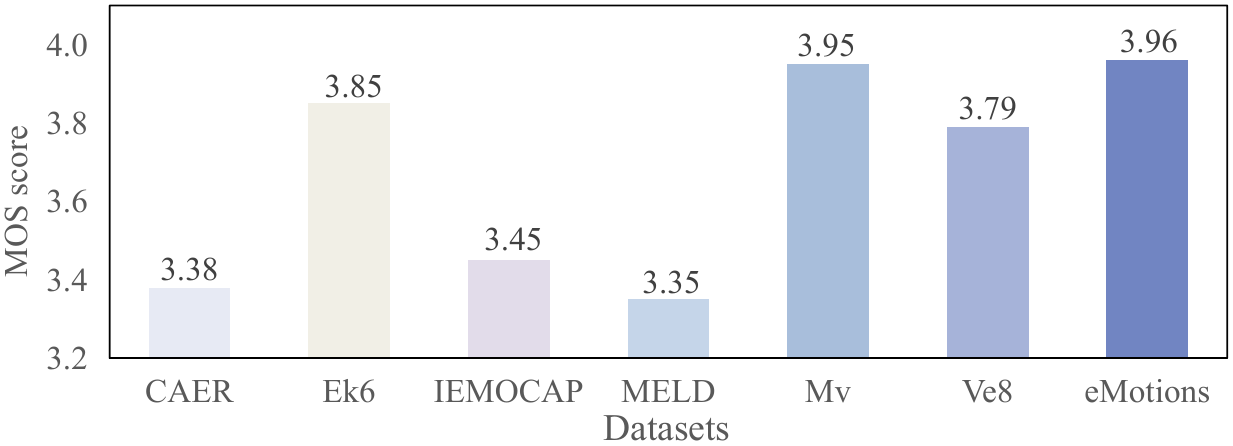} 
\vspace{-2.25em}
\caption{The MOS scores of six VEA datasets and our eMotions. Ek6: Ekman6. Mv: Music\_video. Ve8: VideoEmotion8.}
\label{fig_MOS_Scores}
\vspace{-0.06em}
\end{figure}

\begin{figure*}[]
\setlength{\belowcaptionskip}{-0.1cm}
\centering
\includegraphics[height=3.8cm, width=\textwidth]{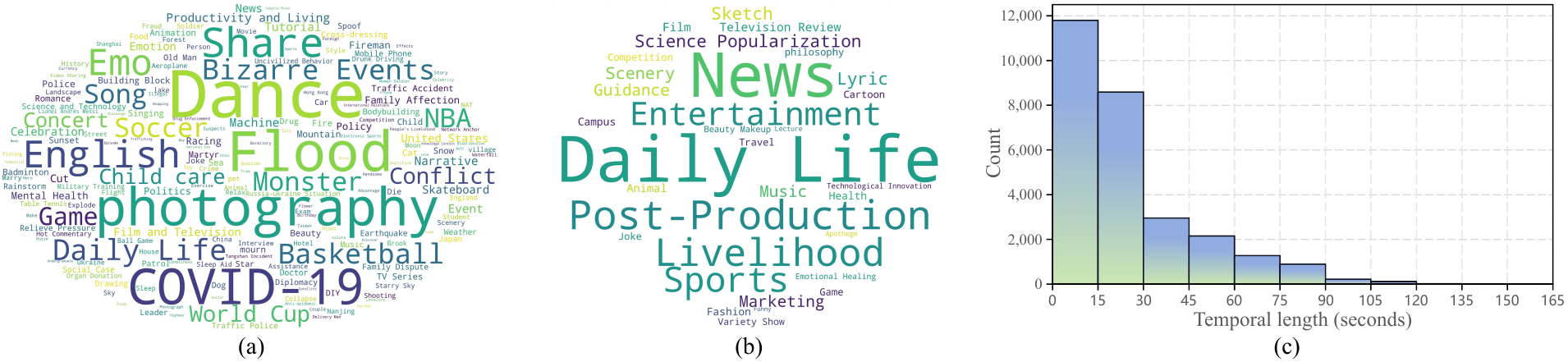}
\vspace{-2.2em}
\caption{(a) \& (b) Word clouds of topics and content types in eMotions. Larger text size indicates a higher frequency of occurrence. (c) Duration distribution of short-form videos in our dataset.}
\label{fig_wordclouds_duration}
\vspace{-0.73em}
\end{figure*}

\subsection{Labeling Quality Evaluations}
\label{subsec:3.4}
In this section, we conduct labeling quality evaluations for eMotions and six VEA datasets. Specifically, we first perform the random sampling of seven datasets, totaling 979 samples, based on the sample correction formula from \cite{new_23_singh2014sampling}. We then employ four emotional annotation \href{https://cloud.baidu.com/product/dcs.html}{experts} to independently evaluate the labeling quality of these datasets. The MOS scores \cite{new_24_ribeiro2011crowdsourcing} are utilized as the evaluation metric, in which the ratings from 1 to 5 represent the five levels of quality (\textit{i.e.}, bad, poor, fair, good, excellent). Next, for each dataset, we average the assessments of four experts and regard the output as the final rating. Although the videos in Music\_video \cite{7_pandeya2021deep} sourced from music recordings are easier to evoke emotions and the inherently unique challenges presented in eMotions, our dataset still ranks the first position, indicating the more reliable annotations compared with existing datasets, as shown in Fig.~\ref{fig_MOS_Scores}. These results verify the effect of our efforts in augmenting labeling quality.

\begin{table}[t!]
\centering
\begin{center}
\setlength{\abovecaptionskip}{-0em} 
\setlength{\arrayrulewidth}{0.22pt}
\caption{The statistics of eMotions, detailing the number of videos and processed frames, along with the quantitative durations. The magnitude is $10^4$ for "Total (s)" and "Frames".}
\label{tab2}
\small
\resizebox{\linewidth}{!}{%
\begin{tabular}{ccccccc}
\toprule[0.22pt]
Category & Videos & \begin{tabular}[c]{@{}c@{}}Total (s)\end{tabular} & \begin{tabular}[c]{@{}c@{}}Shortest (s)\end{tabular} & \begin{tabular}[c]{@{}c@{}}Longest (s)\end{tabular} & \begin{tabular}[c]{@{}c@{}}Average (s)\end{tabular} & \begin{tabular}[c]{@{}c@{}}Frames\end{tabular} \\ \midrule[0.22pt]
Excitation        & 11,739            & 29.35                                                              & 3.72                                                                      & 163.77                                                                   & 25.00                                                                    & 945.29                                                              \\
Fear              & 954              & 2.59                                                               & 2.81                                                                      & 117.49                                                                   & 27.08                                                                    & 78.56                                                               \\
Neutral           & 8,795             & 24.97                                                             & 2.46                                                                      & 150.93                                                                   & 28.39                                                                    & 815.11                                                              \\
Relaxation        & 2,214             & 5.24                                                              & 5.06                                                                      & 117.05                                                                   & 23.69                                                                    & 163.80                                                              \\
Sadness               & 2,090             & 4.04                                                              & 3.25                                                                      & 120.77                                                                   & 19.30                                                                    & 131.53                                                              \\
Tension           & 2,204             & 4.90                                                               & 3.79                                                                      & 119.32                                                                   & 22.25                                                                    & 152.15                                                              \\
\rowcolor{gray!20}
Overall           & 27,996            & 71.09                                                              & 2.46                                                                      & 163.77                                                                   & 25.39                                                                    & 2,286.44                                                             \\ \hline
\end{tabular}%
}
\end{center}
\vspace{-0.5em}
\end{table}

\begin{table}[]
\centering
\setlength{\abovecaptionskip}{-0em} 
\renewcommand{\arraystretch}{0.95}
\setlength{\arrayrulewidth}{0.22pt}
\caption{The quantity and proportions of raw and labeled data across three SVs platforms.}
\label{tab3}
\small
\resizebox{\linewidth}{!}{%
\begin{tabular}{cccccccc}
\toprule[0.22pt]
\multirow{2}{*}{\begin{tabular}[c]{@{}c@{}}Data \\Type\end{tabular}} & \multicolumn{2}{c}{Douyin} & \multicolumn{2}{c}{Kuaishou} & \multicolumn{2}{c}{Tiktok} & \multirow{2}{*}{Sum} \\ \cmidrule[0.22pt](lr){2-7}
                           & No.       & Ratio        & No.         & Ratio         & No.        & Ratio        &                        \\ \midrule[0.22pt]
Raw                   & 15,977       & 47.58\%       & 10,000        & 29.78\%        & 7,600        & 22.64\%       & 33,577                  \\
Labeled              & 12,395       & 44.27\%       & 8,264         & 29.52\%        & 7,337        & 26.21\%       & 27,996                  \\ \bottomrule[0.22pt]
\end{tabular}%
}
\vspace{-1.3em}
\end{table}

\subsection{Dataset Characteristics}
\label{subsec:3.5}
We show the statistics of eMotions in Tab.~\ref{tab2}, including the number of videos and processed frames for each category, as well as the quantitative durations. We observe that Excitation has the largest number of videos, while the negative emotions (\textit{i.e.}, Fear, Sadness, Tension) have the smaller number of videos. In Tab.~\ref{tab3}, we present the quantity and proportions of raw and labeled data across three SVs platforms. We figure out that the videos from Douyin and Kuaishou, two Chinese SVs platforms, account for the largest proportion.

We display the word clouds of topics and content types in Fig.~\ref{fig_wordclouds_duration}~(a) and Fig.~\ref{fig_wordclouds_duration}~(b). It can be seen that the topics mainly focus on daily events in our life, such as dance and photography, as well as the real-time events, such as COVID-19, flood. Content types are  closely connected to human beings (\textit{e.g.}, daily life, livelihood, news). Moreover, some topics and content types of SVs are consistent (\textit{e.g.}, game, health). Fig.~\ref{fig_wordclouds_duration}~(c) presents the distribution of SVs duration.  We find that the durations mainly concentrate from 0 to 30 seconds, taking up 72.81\% of the overall dataset. This indicates that SVs are concise, which can meet the needs of online users for quick dissemination of web information. Besides, different from the human-centered VEA datasets (\textit{e.g.}, \cite{16_zadeh2018multimodal,26_busso2008iemocap,33_soujanya2018multimodal}), eMotions originating from SVs is directed towards in-the-wild scenes, implying that the emotional elicitation requires the whole duration, not short clips of video.

\subsection{Definitions of Two Variant Datasets}
\label{subsec:3.6}
We build two variant datasets through targeted selections to meet various research needs. For eMotions\_balanced, we randomly select samples from eMotions, in which 4,000 samples of Excitation, 3,000 samples of Neutral, and all the samples of the remaining categories. Regarding eMotions\_test, we randomly select the corresponding proportion of samples based on the ratio of each emotional category in eMotions, totaling 5,000 samples. We detailedly show the statistics of two variant datasets in the appendix.

\section{Methodology}
\label{sec:method}

\begin{figure*}[t!]
\setlength{\belowcaptionskip}{-1.5em}
\centering
\includegraphics[height=4.1cm,width=\textwidth]{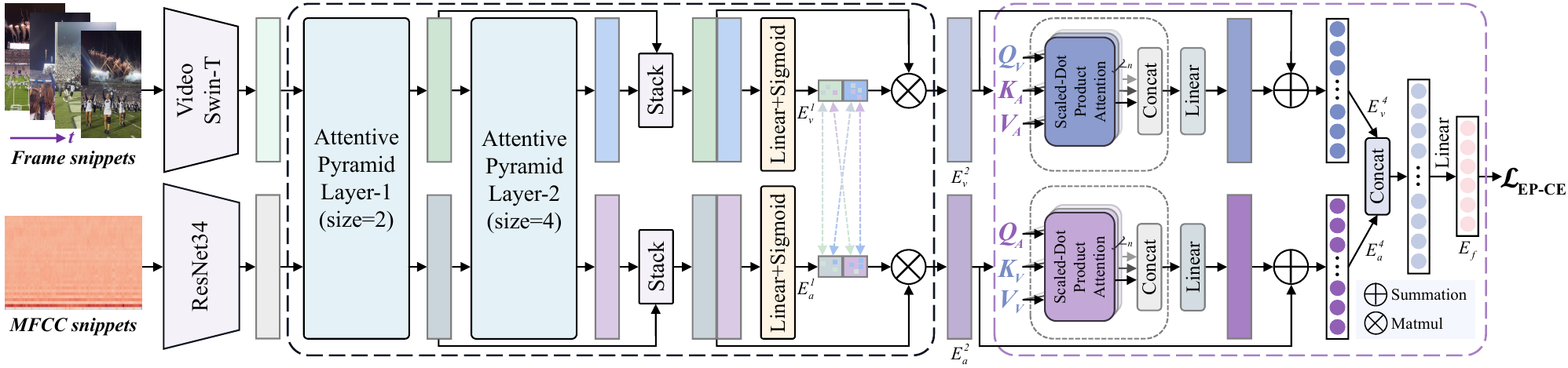}
\vspace{-2.05em}
\caption{The overall illustration of AV-CANet. The Local-Global Fusion (LGF) Module consists of Local-level Interactive-Selective Fusion and Global-level Complementary Fusion Sub-Modules, framed by dashed lines in \textit{black} and \textit{\textcolor[RGB]{167,127,197}{purple}}, respectively.}
\label{fig5-AV-CANet}
\end{figure*}

\subsection{Visual and Audio Representations}
\label{subsec:4.1}
To tackle the inherent challenges of eMotions, we end-to-end learn audio-visual representations. Unlike \cite{35_zhao2020end,40_zhang2023weakly}, we employ the Video Swin-Transformer (Video Swin-T) \cite{19_liu2022video} as the visual backbone since it can capture more semantically relevant features. Specifically, let $\{(v_i, a_i)\}^b$ be a batch of $b$ samples, in which $v_i$, $a_i$ denote the video and audio in $i$-th sample. For $v_i$, we divide it into $s$ segments of equal duration and randomly select $T$ successive frames from each segment by keeping the temporal order, as the input snippets. For each snippet-level input $ v_i^{j} \in \mathbb{R}^{T \times H \times W \times 3}$, we first project it into the patch features $ v_i^{j}\in \mathbb{R}^{\frac{T}{2} \times \frac{H}{4} \times \frac{W}{4} \times 96}$, where $H$ and $W$ refer to the height and width of $v_i$. For the $l$-th stage in Video Swin-T, the input features can be represented as $F_v^{l} = \frac{T}{2} \times \frac{H}{2^{l+1}} \times \frac{W}{2^{l+1}} \times 2^{l-1}C $, $l \in \{1, 2, 3, 4\}$. For the first stage, a linear embedding layer is applied to project the dimension of patch features to $C$. The patch merging layers are then utilized in the remaining stages to progressively reduce the spatial size of feature maps. Moreover, the blocks in Video Swin-T deploy the novel 3D SW-MSA Module to capture global relations and model spatial-temporal correlations. The final representations of $v_i$ comprise a series of snippet features $F_{v}(v_i) = \{f_{v}^{1}(v_i), f_{v}^{2}(v_i), ..., f_{v}^{s}(v_i)$\}, and $f_{v}^{j}(v_i) \in \mathbb{R}^{\frac{T}{2} \times \frac{H}{32} \times \frac{W}{32} \times 8C}$.

For audio $a_i$, we obtain a successive descriptor $a$ through MFCC. We then center-crop $a$ to a fixed length of $q$ and pad itself when necessary to get $a^\prime$. Next, we divide $a^\prime$ along the time series length into $s$ snippets via chunking and stacking, utilize ResNet34 \cite{22_he2016deep} to output the final representations $F_{a}(a_i) = \{f_{a}^{1}(a_i), f_{a}^{2}(a_i), ..., f_{a}^{s}(a_i)\}$, and $ f_{a}^{p}(a_i) \in \mathbb{R}^{ H^\prime \times W^\prime \times C^\prime}$, in which $H^\prime$, $W^\prime$, and $C^\prime$ denote the height, width, and final dimension, respectively.

Afterwards, we deploy poolings along spatial-temporal dimensions, followed by the fully connected layers to reshape $F_{v}$ and $F_{a}$ as $ F_{v} \in \mathbb{R}^{s \times C_{1}}$ and $ F_{a} \in \mathbb{R}^{s \times C_{1}}$.

\subsection{Local-Global Fusion Module}
\label{subsec:4.2-new}
We propose the LGF Module, including Local-level Interactive-Selective Fusion and Global-level Complementary Fusion (\textit{i.e.}, LISF and GLCF) Sub-Modules, to progressively model the correlations of inter-modalities, thereby mitigating the local biases and collective information gaps caused by emotion inconsistence.

In the LISF Sub-Module, we first serially adopt two attentive pyramid layers \cite{new_37_yu2022mm}, building upon the reformed Self-Attention (SA) and Cross-Modal Attention (CMA) blocks, to provide locally dense interactions of audio-visual features. Concretely, to facilitate interactions at multiple contextual scales in the pyramid layers, we set the fixed interaction window $s_t(F,d) = [F_{t-d}, ..., F_{t+d}]$ by adding masks to regions that should not be involved, where $d$ denotes the size for $t^{th}$ snippet and $t \in [1, s]$, constraining the interacting size of blocks. With this fixed-size mechanism, the attention calculation of SA and CMA blocks at snippet-level can be formulated as:
\begin{align}
SA(F_{m},d) &= Att({F_{m}}W_{q}, s_t(F_{m},d)W_{k},  s_t(F_{m},d)W_{v}), \\
CMA(F_{m}, F_{\bar{m}}, d) &= Att({F_{m}}W_{q}, s_t(F_{\bar{m}},d)W_{k},  s_t(F_{\bar{m}},d)W_{v}), \\
Att(q, k, v) &= softmax(qk^T/\sqrt{d_k})v,
\label{eq-sa-cma-att}
\end{align}
where $q$, $k$, $v$, and $1/\sqrt{d_k}$ refer to queries, keys, values, and the scaling factor, $W_{*}$ ($* \in \{q,k,v \}$) are learnable parameters, modality $ m \in \{a, v\}$. Subsequently, the reformed feed-forward layer, LayerNorm~\cite{new_50_ba2016layer}, and residual connection are deployed to form the complete SA and CMA blocks.

When multiple attention heads are adopted, the outputs of SA and CMA blocks can be represented as:
\begin{align}
F_{s}^m &= \mathbf{c}\left(Att_{sa}^1, ..., Att_{sa}^n\right)W_0, \\
F_{c}^m &= \mathbf{c}\left(Att_{cma}^1, ..., Att_{cma}^n\right)W_1,
\label{eq-sa-cma-mh}
\end{align}
where $\mathbf{c}(\cdot)$ and $n$ denote concatenation and the number of attention heads, $W_0$ and $W_1$ are learnable parameters. Besides, the parameters in CMA block are shared to facilitate local-level interactions.

For the parallel outputs of SA and CMA blocks, we first concatenate them and compute channel-wise attention scores to perform refinement via a linear layer followed by a sigmoid function, then summation is used to output the local-aware audio-visual features:
\begin{equation}
\small
F_m^\prime = \sigma\left(W_{s} \cdot \mathbf{c}\left(F_{s}^m, F_{c}^m\right) + b_s\right)F_{s}^m + \sigma\left(W_{c} \cdot \mathbf{c}\left(F_{s}^m, F_{c}^m\right) + b_c\right)F_{c}^m,
\label{eq-overall}
\end{equation}
where $\mathbf{\sigma}(\cdot)$ denotes sigmoid function, $W_{*}$ and $b_{*}$ ($* \in \{s,c \}$) are the learnable parameters of linear layers.

Following \cite{new_37_yu2022mm,new_38_li2020ms}, the acausally dilated residual blocks are then adopted to amalgamatedly derive temporal semantics and perform positional enhancements. In the end, the densely interactive outputs $\{F_v^i, F_a^i\}_{i=1}^{L}$ for $t^{th}$ snippet are preserved as pyramid-like features, in which $ F_m^i \in \mathbb{R}^{1 \times C_{1}}$ and $L$ is the number of pyramid layers.

\begin{table*}[t]
\centering
\begin{center}
\setlength{\abovecaptionskip}{-0em} 
\setlength{\arrayrulewidth}{0.22pt}
\caption{Performace comparisons of recent VEA methods and our AV-CANet on three eMotions-related datasets. We report the overall metrics of Accuracy (ACC) and Weighted Average F1-Score (WA-F1), as well as the ACC of each emotional category. EX: Excitation. FR: Fear. NU: Neutral. RE: Relaxation. SD: Sadness. TN: Tension. e\_B: eMotions\_balanced. e\_T: eMotions\_test. e\_All: eMotions. Throughout this paper, we highlight the best performance in \textbf{\underline{\textit{bold}}} and \underline{underline} the second performance.}
\label{tab-performace-comparisons}

\rule{\linewidth}{0pt}
{
\renewcommand{\arraystretch}{1.350} 
\tabcolsep=0.6136pt
\fontsize{6.55}{7.2}\selectfont
\begin{tabular}{lclcccccccccccccccccclcccccc}
\hline
\multicolumn{2}{c}{\hspace{0.32cm}\multirow{3}{*}{Method}}               &  & \multicolumn{18}{c}{ACC of Each Emotional Category (\%)}                                                                                                  &  & \multicolumn{6}{c}{Evaluation Metrics (\%)}                     \\ \cline{4-21} \cline{23-28} 
\multicolumn{2}{c}{}                                      &  & \multicolumn{3}{c}{EX} & \multicolumn{3}{c}{FR} & \multicolumn{3}{c}{NU} & \multicolumn{3}{c}{RE} & \multicolumn{3}{c}{SD} & \multicolumn{3}{c}{TN} &  & \multicolumn{3}{c}{ACC} & \multicolumn{3}{c}{WA-F1} \\ \cline{4-21} \cline{23-28} 
\multicolumn{2}{c}{}                                      &  & e\_B  & e\_T  & e\_All & e\_B  & e\_T  & e\_All & e\_B  & e\_T  & e\_All & e\_B  & e\_T  & e\_All & e\_B  & e\_T  & e\_All & e\_B  & e\_T  & e\_All &  & e\_B   & e\_T  & e\_All & e\_B   & e\_T   & e\_All  \\ \hline
\multirow{5}{*}{A}   & ResNet34  \cite{22_he2016deep}                           &  & 66.63 & 70.00 & 78.92  & 50.79 & 41.18 & 37.17  & 37.67 & 57.32 & 55.66  & 68.62 & 51.90 & 44.70  & 37.08 & 12.00 & 16.27  & 52.38 & 17.95 & 22.90  &  & 53.44  & 55.20 & 58.39  & 52.65  & 53.13  & 56.14   \\
                     & PANNS \cite{27_kong2020panns}                              &  & 71.25 & 78.10 & \underline{83.48}  & 49.74 & 41.18 & 43.46  & 30.17 & 45.86 & 42.58  & 57.79 & 41.77 & 50.79  & 36.60 & 4.00  & 16.75  & 51.02 & 5.13  & 29.93  &  & 51.16  & 52.60 & 57.48  & 50.06  & 48.48  & 54.99   \\
                     & ECAPA-TDNN \cite{new_32_desplanques2020ecapa}                         &  & 69.13 & \underline{80.24} & \underline{\textbf{84.63}}  & 42.93 & 29.41 & 29.32  & 36.83 & 39.81 & 47.24  & 49.21 & 31.65 & 42.66  & 55.02 & 36.00 & 23.68  & 38.78 & 14.10 & 13.38  &  & 50.99  & 53.50 & 57.52  & 50.45  & 50.92  & 54.55   \\
                     & Res2Net \cite{34_gao2019res2net}                            &  & \underline{\textbf{73.63}} & 75.24 & 71.98  & 43.98 & 32.35 & 40.84  & 35.83 & 47.13 & 51.51  & 57.79 & 45.57 & 48.98  & 26.08 & 5.33  & 17.70  & 44.44 & 20.51 & 30.39  &  & 50.09  & 53.10 & 55.34  & 48.72  & 50.37  & 53.96   \\
                     & CAMPPlus \cite{38_wang2023cam++}                           &  & 69.00 & 73.03 & 78.62  & 40.31 & 41.18 & 34.56  & 36.50 & 48.09 & 49.91  & 65.46 & 51.90 & 48.08  & 38.28 & 10.67 & 13.88  & 48.53 & 25.32 & 18.14  &  & 52.26  & 54.00 & 56.09  & 51.37  & 52.06  & 53.51   \\ \hline
\multirow{8}{*}{V}   & Video Swin-Transformer \cite{19_liu2022video}             &  & 66.67 & 66.83 & 65.64  & 66.14 & \underline{61.76} & 49.21  & \underline{\textbf{69.76}} & \underline{80.25} & \underline{\textbf{86.80}}  & 71.37 & 56.96 & 55.30  & 51.63 & 40.00 & 40.19  & 55.03 & 11.39 & 10.20  &  & 64.12  & 63.70 & 64.64  & \underline{64.31}  & 62.21  & 62.92   \\
                     & I3D \cite{43_carreira2017quo}                                &  & 69.84 & 68.74 & 71.48  & 74.87 & \underline{\textbf{64.71}} & 52.36  & 62.83 & \underline{\textbf{80.89}} & 80.66  & 70.20 & \underline{\textbf{59.49}} & 51.92  & 51.67 & 37.33 & 42.58  & 51.02 & 6.33  & 13.15  &  & 63.28  & \underline{64.40} & 65.41  & 63.68  & \underline{62.68}  & 63.90   \\
                     & SlowFast \cite{new_33_feichtenhofer2019slowfast}                           &  & 70.34 & 76.37 & 77.71  & 5.76  & 0.00  & 0.00   & 22.33 & 47.77 & 51.42  & 38.37 & 7.59  & 15.35  & 45.93 & 14.67 & 11.24  & 48.98 & 0.00  & 3.40   &  & 44.43  & 48.70 & 51.05  & 41.94  & 42.77  & 45.60   \\
                     & CTEN \cite{40_zhang2023weakly}                              &  & 65.00   & 74.22 &73.51        &37.17       & 38.24 &28.80        &45.83       & 61.46 &65.87        &69.07       & 54.43 &45.82        &52.15       & 32.00 &30.62        &36.28       & 7.59  &6.12        &  &53.58        & 59.00 &58.89        &53.02        & 56.94  &56.53         \\
                     & Former-DFER \cite{new_34_zhao2021former}                        &  & 64.85 & 74.46 & 69.97  & 41.80 & 38.24 & 39.79  & 36.01 & 50.32 & 65.55  & 58.57 & 26.58 & 51.69  & 43.36 & 24.00 & 39.71  & 51.68 & 26.58 & 10.43  &  & 51.64  & 54.30 & 59.16  & 51.38  & 52.56  & 57.67   \\
                     & TimeSformer \cite{30_bertasius2021space}                        &  & 68.84 & 70.88 & 78.90  & 74.35 & 55.88 & 51.31  & \underline{63.33} & 75.48 & 73.09  & 73.81 & 40.51 & 47.18  & 57.89 & 28.00 & 34.45  & 48.75 & 8.86  & 4.31   &  & \underline{64.18}  & 61.30 & 64.43  & \underline{\textbf{64.32}}  & 59.17  & 61.62   \\
                     & DFER-CLIP \cite{new_36_zhao2023prompting}                             &  & 59.40 & 67.48 & 69.55   & \underline{\textbf{84.29}} & 50.00 & \underline{\textbf{92.38}}   & 45.49  & 53.44  & 57.12   & \underline{74.32}  & 57.14  & \underline{\textbf{73.56}}   & 63.98  & \underline{44.68}  & \underline{\textbf{67.60}}   & 51.42  & \underline{39.13}  & \underline{40.80}   &  & 58.82   & 58.90  & 64.33   & 59.04   & 56.96  & 63.07    \\
                     & C3D \cite{32_tran2015learning}                                &  & \underline{72.84} & 76.61 & 65.94  & 69.63 & 55.88 & 41.88  & 51.33 & 62.74 & \underline{82.76}  & 69.53 & 49.37 & 58.92  & 55.50 & 37.33 & 43.78  & 51.25 & 12.66 & 3.40   &  & 61.86  & 61.40 & 63.27  & 61.84  & 59.72  & 61.04   \\ \hline
\multirow{6}{*}{A+V} & VAANet \cite{35_zhao2020end}                             &  & 72.38 & 
                     74.46 & 72.79  & 63.35 & 47.06 & 58.64  & 47.83 & 57.01 & 65.43  & 67.95 & 54.43 & 59.14  & 44.50 & 28.00 & 45.22  & 59.41 & 37.97 & 32.88  &  & 60.01  & 60.10 & 63.71  & 59.89  & 59.46  & 63.31   \\
                     & OGM-GE \cite{50_peng2022balanced}                             &  & 67.27 & 77.00 & 80.32  & 67.20 & 35.29 & 34.34  & 33.92 & 48.30 & 42.71  & 46.64 & 27.69 & 39.62  & 31.33 & 26.83 & 18.33  & 56.60 & 8.57  & 34.20  &  & 50.78  & 54.20 & 55.39  & 50.02  & 51.49  & 53.19   \\
                     & MSAF \cite{new_35_su2020msaf}                             &  & 60.33  & 63.43  & 71.57   & 54.77  & 44.83  & 51.32   & 41.30  & 60.52  & 65.06   & 70.95  & 37.18  & 60.92   & 52.36  & 26.39  & 37.03   & 47.80  & 15.28   & 12.33   &  & 54.69   & 53.73  & 60.93   & 54.51   & 52.70   & 59.51    \\
                     & MMCosine \cite{53_xu2023mmcosine}                          &  & 59.03 & 78.17 & 68.93  & 35.45 & 23.53 & 48.99  & 41.08 & 45.51 & 53.87  & 59.22 & 30.77 & 43.63  & 57.89 & 28.05 & 34.29  & 50.34 & 7.14  & 23.53  &  & 52.47  & 53.60 & 55.27  & 52.46  & 50.60  & 54.35   \\
                     & 3D-ResNet50 + ResNet34 \cite{21_hara2018can,22_he2016deep}             &  & 72.50 & 68.02 & 77.73  & 69.63 & 35.29 & 60.21  & 49.17 & 69.43 & 59.24  & 72.69 & \underline{58.23} & 60.50  & 55.74 & 36.00 & 36.36  & 55.78 & 17.72 & \underline{\textbf{41.04}}  &  & 62.53  & 60.20 & 63.98  & 62.31  & 59.04  & 63.45   \\
                     & ConvNeXt + CSP-DarkNet53 \cite{46_liu2022convnet,42_wang2020cspnet} &  & 66.25 & 77.33 & 78.36  & \underline{75.92} & 50.00 & \underline{66.49}  & 46.17 & 62.74 & 61.46  & 69.30 & 50.63 & 57.11  & \underline{\textbf{65.55}} & 34.67 & \underline{54.55}  & \underline{\textbf{65.08}} & 35.44 & 33.11  &  & 62.91  & 63.20 & \underline{65.62}  & 62.87  & 62.33  & \underline{65.01}   \\
                     \rowcolor{gray!20}
                     & AV-CANet                           &  & 67.03   & \underline{\textbf{82.58}}   & 80.24    & 68.25   & 58.82   & 59.16    & 50.87   &62.10       & 63.39        & \underline{\textbf{75.05}}       & 45.57  & \underline{63.43}        & \underline{65.16}   & \underline{\textbf{48.00}}       & 54.07       & \underline{63.53}       & \underline{\textbf{46.84}}   & 40.14    &  & \underline{\textbf{64.40}}        & \underline{\textbf{67.00}}       & \underline{\textbf{67.79}}        & \underline{\textbf{64.32}}        & \underline{\textbf{66.28}}        & \underline{\textbf{67.32}}         \\ \hline
\end{tabular}}
\end{center}
\vspace{-0.95em}
\end{table*}

To adaptively conduct selective integrations of pyramid features at snippet-level, we propose the Selectively Cross-Aggregative Integration Layer instead of simply using poolings. To be specific, for $t^{th}$ snippet, we first stack $\{F_m^i\}_{i=1}^{L}$ along $L$ to unify interactive features at multiple pyramid layers. The linear projection and sigmoid function are then performed to dynamically assign weights for different granularities, which can be formulated as:

\vspace{-0.7em}
\begin{equation}
E_m^1 = \sigma\left(W_a \cdot s\left(F_m^1, ..., F_m^L\right)+b_a\right),
\end{equation}
where $s(\cdot)$ refers to stacking operation. Afterwards, we deploy the CMA block without fixed-size window, where the number of head and dimension both are 1, to adaptively adjust the importance of audio-visual weights. In the end, the weighted summation of pyramid layers is conducted to output the compatibly integrated features $E_m^2$. The formulation of above operations is given as follows:
\vspace{-0.7em}
\begin{equation}
E_m^2 = \sum_{l=1}^{L} s^{l}\left(F_m^1, ..., F_m^L\right) \cdot Att^{l}\left(E_m^1 W_q, E_{\bar{m}}^1 W_k, E_{\bar{m}}^1 W_v\right).
\end{equation}

So far, our model learns representations that are well-aware of local biases. To mitigate the collective information gaps, we introduce the GLCF Sub-Module, which facilitates correlated information capture of audio-visual features at global-level and outputs the final results. Specifically, we first utilize the globally unrestricted CMA block \cite{57_vaswani2017attention} with multiple attention heads to complementarily learn audio-visual representations at various levels and different degrees of correlations between snippets $s$. Then, concatenation is performed for non-local aggregations, \textit{i.e.},

\vspace{-0.7em}
\begin{equation}
\small
E_m^3 = \mathbf{c}\left(Att_{gca}^1, ..., Att_{gca}^i\left(E_m^2 W_q, E_{\bar{m}}^2 W_k, E_{\bar{m}}^2 W_v\right), ...,Att_{gca}^n\right) W_2.
\end{equation}

Afterwards, the linear projection layers are deployed to enhance the focus on discriminative information and snippet-aware correlations. The residual connection is also utilized to benefit the free information flow and gradient propagation, followed by pooling layers along the snippet dimension to reshape features as $E_m^4 \in \mathbb{R}^{C_{2}}$:
\begin{equation}
\vspace{-0.12em}
E_m^4 = \delta\left(\left(W_{g} \cdot E_m^3 + b_{g}\right)+E_m^2\right),
\end{equation}
where $\delta(\cdot) $ denotes average pooling. In the end, we adopt the Mid-Concat fusion strategy, followed by a category-specific linear layer to output the final representations of our overall model, \textit{i.e.},
\begin{equation}
\vspace{-0.12em}
E_f = W_{f} \cdot \mathbf{c}\left(E_m^4, E_{\bar{m}}^4\right) + b_{f}.
\end{equation}

\subsection{Emotion Polarity Enhanced CE Loss} 
\label{subsec:4.3} 
Directly optimizing the traditional CE Loss could lead to misclassification due to the more significant semantic gaps and difficulties in learning emotion-related features. Inspired by \cite{35_zhao2020end}, we develop the EP-CE Loss, which further considers the \textit{neutral} polarity presented in eMotions and many VEA datasets \cite{7_pandeya2021deep,18_lee2019context,26_busso2008iemocap}. Specifically, when the emotion polarity of prediction is different from the ground-truth, the $\gamma_{\textit{ep}(y_i)}$ will function to weight model optimization. The EP-CE Loss can be defined as:
\vspace{-0.25em}
\begin{equation}
\mathcal{L}_{ep} = -\frac{1}{N} \sum_{i=1}^N \left(1 + \gamma_{\textit{ep}(y_i)} \cdot s(y_i, \hat{y_i})\right) \sum_{c=0}^{C-1} \beta_{[c=y_i]} \log p_{i, c},
\label{eq11}
\vspace{-0.4em}
\end{equation}
where $C$ is the number of categories, $\beta_{\left[c=y_i\right]}$ is a binary indicator, and $p_{i, c}$ is the predicted probability that sample $i$ belongs to category $c$. $\gamma_{\textit{ep}(y_i)}$ refers to the coefficients for polarities that control the penalty extents. $y_i$ and $\hat{y_i}$ refer to the ground-truth and prediction. $s\left(y_i, \hat{y_i}\right)$ represents whether to add the penalties. When $\operatorname{\textit{ep}}(y_i) \neq \operatorname{\textit{ep}}(\hat{y_i})$, $s\left(y_i, \hat{y_i}\right) = 1$, otherwise $s\left(y_i, \hat{y_i}\right) = 0$, where \textit{ep}$(\cdot)$ maps the emotional category to its corresponding polarity. 

Furthermore, to improve the training stability on large-scale dataset, we first deploy the deferred update strategy after every 4 batches, then adopt the multi-task learning manner, which accumulates the loss of overall, visual and audio branches.

\begin{table}[]
\centering
\setlength{\abovecaptionskip}{-0em} 
\renewcommand{\arraystretch}{0.99} 
\setlength{\arrayrulewidth}{0.22pt}
\caption{Performance comparisons of our LGF Module with advanced overall fusion methods.}
\small
\resizebox{\linewidth}{!}{%
\begin{tabular}{ccccccc}
\toprule[0.22pt]
\multirow{2}{*}{\raisebox{-2ex}{\begin{tabular}[c]{@{}c@{}}Method\end{tabular}}} & \multicolumn{2}{c}{eMotions}    & \multicolumn{2}{c}{Music\_video} & \multicolumn{2}{c}{Ekman6}      \\ \cmidrule[0.22pt](l){2-3}   \cmidrule[0.22pt](l){4-5}   \cmidrule[0.22pt](l){6-7} 
                                                                            & ACC         & WA-F1          & ACC         & WA-F1          & ACC         & WA-F1          \\ \midrule[0.22pt]
Vanilla                                                                & 65.98          & 65.05          & 75.19          & 74.66 & 50.31          & 50.41          \\
Self-Attention \cite{57_vaswani2017attention}                                                                & \underline{67.21}          & \underline{66.72}          & 81.77          & 81.56 & 51.56          & 50.73          \\
LGI-Former \cite{new_52_sun2023mae}          & 67.13          & 66.51          & \underline{82.03}          & \underline{82.08}          & \underline{53.75}          & \underline{53.20}          \\
Sun et al. \cite{new_45_sun2024hicmae}                                                         & 66.14          & 65.15          & 77.97          & 77.62          & 49.69         & 49.29          \\
\rowcolor{gray!20}
LGF Module                                                                   & \textbf{\underline{67.48}} & \textbf{\underline{67.19}} & \textbf{\underline{82.78}} & \textbf{\underline{82.46}}          & \textbf{\underline{54.06}} & \textbf{\underline{53.51}} \\ 
\hline
\end{tabular}
}
\label{tab-fusion-compare}
\vspace{-1.8em}
\end{table}

\begin{table*}[!ht]
\centering
\setlength{\abovecaptionskip}{-0em}
\setlength{\arrayrulewidth}{0.22pt}
\caption{The overall performace comparisons of recent baseline methods and AV-CANet on Ekman6 (Ek6), VideoEmotion8 (Ve8), Music\_video (Mv), and IEMOCAP (IE). EM: Evaluation metrics. *: The lab-controlled dataset. -: The results are not available.}
\label{tab-performance-comparisons-public-datasets}
\noindent
\rule{\linewidth}{0pt}
{
\small
\renewcommand{\arraystretch}{1.25}
\tabcolsep=1.67pt
\begin{tabular}{cclccccclccccccclcccccccclc}
\hline
\multirow{2}{*}{Dataset} & \multirow{2}{*}{EM} &  & \multicolumn{5}{c}{A}                                &  & \multicolumn{8}{c}{V}                                                    &  & \multicolumn{8}{c}{A+V}                                                                      \\ \cline{4-8} \cline{10-17} \cline{19-27}
                         &                         &  & \cite{22_he2016deep} & \cite{27_kong2020panns} & \cite{new_32_desplanques2020ecapa} & \cite{34_gao2019res2net}  & \cite{38_wang2023cam++} &  & \cite{19_liu2022video} & \cite{43_carreira2017quo} & \cite{new_33_feichtenhofer2019slowfast} & \cite{40_zhang2023weakly}  & \cite{new_34_zhao2021former} & \cite{30_bertasius2021space} & \cite{new_36_zhao2023prompting} & \hspace{0.05cm}\cite{32_tran2015learning} &   & \cite{35_zhao2020end} & \cite{50_peng2022balanced} & \cite{7_pandeya2021deep} & \cite{new_35_su2020msaf} & \cite{53_xu2023mmcosine} & \cite{21_hara2018can,22_he2016deep} & \cite{46_liu2022convnet,42_wang2020cspnet} &  & \textbf{Ours}             \\ \hline
\multirow{2}{*}{Ek6}     & ACC                     &  & 33.95    & 27.47    & 28.40    & 31.17    & 29.01    &  & 47.59    & 40.51   & 28.35    & \underline{53.44}    & 37.96    & 36.33   & 44.44 & 44.05    &   & 50.31    & 34.26    & -      & 30.56      & 33.33    & 48.44       & 50.00       &  & \underline{\textbf{55.63}}              \\
                         & WA-F1                   &  & 32.17    & 21.77    & 28.21    & 29.64    & 29.03    &  & 49.44    & 42.10   & 21.92    & \underline{52.76}    & 36.88    & 34.34   & 43.81 & 42.84    &  & 49.65    & 33.10    & -      & 25.35      & 33.08    & 47.57       & 49.33       &  & \underline{\textbf{55.44}}              \\ \hline
\multirow{2}{*}{Ve8}     & ACC                     &  & 31.63    & 27.91    & 26.98    & 29.77    & 22.33    &  & 46.31    & 40.89   & 25.12    & \underline{47.66}    & 35.35    & 46.30   & 37.21 & 42.36    &   & 46.73    & 33.33    & -      & 31.25      & 32.18    & 44.39       & 45.33       &  & \underline{\textbf{49.53}}                       \\
                         & WA-F1                   &  & 29.11    & 23.06    & 23.20    & 27.12    & 23.59    &  & 45.77    & 41.34   & 16.06    & \underline{47.04}    & 29.10    & 45.65   & 29.32 & 42.02    &  & 43.92    & 32.12    & -      & 28.56      & 26.82    & 43.21       & 45.23       &  & \underline{\textbf{48.24}}                   \\ \hline
\multirow{2}{*}{Mv}      & ACC                     &  & 72.73    & 63.38    & 69.95    & 65.91    & 70.71    &  & 74.68    & 64.05   & 50.13    & 76.20    & 60.51    & 78.23   & 79.49 & 62.78    &  & 78.99    & 57.32    & \underline{83.30}    & 63.38      & 58.59    & 73.42       & 75.70       &  & \underline{\textbf{84.81}}                    \\
                         & WA-F1                   &  & 73.07    & 61.46    & 69.95    & 65.76    & 71.79    &  & 80.64    & 79.75   & 45.17    & 75.43    & 59.73    & 81.07   & 79.51 & 77.43    &  & 77.99    & 56.46    & \underline{84.00}    & 63.06      & 57.70    & 72.98       & 74.25       &  & \underline{\textbf{84.63}}                    \\ \hline
\multirow{2}{*}{IE*}    & ACC                     &  & 53.09    & 53.44    & 55.31    & 48.78    & 51.81    &  & 68.48    & 68.64   & 27.54    & 55.31    & 54.26    & 38.27   & 32.91 & \underline{\textbf{68.96}}    &  & 62.08    & 49.07    & -      & 50.70      & 49.42    & 63.36       & 61.03       &  & \underline{68.73}                    \\
                         & WA-F1                   &  & 52.45    & 52.83    & 55.17    & 47.36    & 50.88    &  & 67.44    & 67.99   & 16.67    & 54.96    & 54.18    & 31.81   & 16.29 & \underline{68.16}    &  & 61.86    & 48.82    & -      & 49.76      & 49.42    & 63.12       & 60.26       &  & \underline{\textbf{68.73}}                    \\ \hline
\end{tabular}
}
\vspace{-0.68em}
\end{table*}

\section{Experiments}
\label{sec:experiment}
\subsection{Main Results}
\textbf{Performance Comparisons.} To demonstrate the effectiveness of proposed model, we compare with 19 baseline methods on three eMotions-related datasets, in which the modalities of compared baselines include audio (A), visual (V), and audio-visual (A+V), as illustrated in Tab.~\ref{tab-performace-comparisons}. We draw the following observations: (1) Visual features dominate audio-visual VEA, and the stronger visual architecture generally leads to more performance improvements. (2) Compared with other methods, AV-CANet exhibits relatively balanced performance arcoss six emotions on three datasets, indicating that it can adeptly understand the emotions in SVs. (3) Benefit from the effective design of tackling challenges of eMotions, AV-CANet achieves superior performance across the overall metrics, exhibiting the improvements of 4.08\% ACC and 4.01 WA-F1 on eMotions compared to VAANet~\cite{35_zhao2020end}. Besides, we present the comparison results in terms of UAR and WAR across three datasets in the provided appendix.

In Tab.~\ref{tab-performance-comparisons-public-datasets}, the comparative results on four public VEA datasets show that our model performs favorably against recent VEA methods of different modalities, demonstrating its generalizability and robustness for various application-oriented VEA. Additionally, considering the observations in model performance across datasets, some VEA-oriented methods underperform the fine-tuned general models with large-sacle pre-training, implying the importance of large-scale datasets for VEA \cite{new_45_sun2024hicmae,new_52_sun2023mae}. Note that in the appendix, we present more analysis and qualitative evaluations.

\begin{figure}[]
\setlength{\belowcaptionskip}{-0.5cm}
\centering
\includegraphics[height=2.65cm, width=\linewidth]{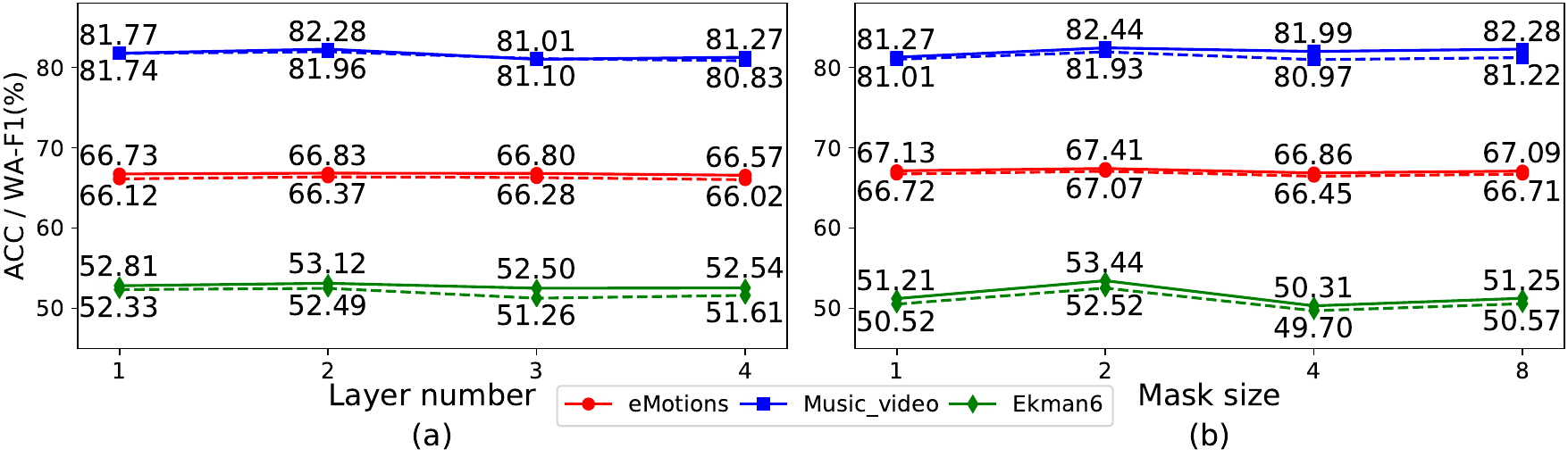}
\vspace{-2.28em}
\caption{Ablation study in investigating the influence of different layer number and mask sizes on LGF Module. The solid and dashed lines denote ACC and WA-F1, respectively.}
\label{fig-hryper}
\vspace{0.2604em}
\end{figure}

\noindent \textbf{The Effectiveness of LGF Module.} As shown in Tab.~\ref{tab-fusion-compare}, we fairly compare our LGF Module with advanced overall fusion methods using the same backbones on three datasets. We figure out that the LGF Module consistently outperforms the advanced methods, which exceeds the second performance up to 0.75\% ACC and 0.38 WA-F1, indicating its effectiveness. Besides, we replace the fusion modules of five audio-visual baselines with the LGF Module to further investigate its impact, as shown in the appendix.

\subsection{Ablation Studies}
We systematically conduct in-depth ablation studies to investigate the key factors of AV-CANet on three datasets. \textbf{(1)} We first examine the influence of different layer number in the LISF Sub-Module, as illustrated in Fig.~\ref{fig-hryper} (a). We conclude that the number of pyramid layers is not directly proportional to the gains, as too many layers result in overly dense interactions at local-level, leading to increased complexity and instability. \textbf{(2)} Next, we evaluate how different mask sizes affect performance, as displayed in  Fig.~\ref{fig-hryper} (b). We observe that inappropriate mask sizes lead to performance declines, indicating that our model are sensitive to mask size. \textbf{(3)} We then investigate the impact of different fusion strategies in the GLCF Sub-Module of LGF Module. As shown in Tab.~\ref{tab-fusion-strategies}, adopting Mid-Concat fusion strategy across three datasets achieves the largest performance improvement of 0.94\% Acc and 1.02 WA-F1, while EW-Multiply and Sum fusion strategies result in performance declines across all the datasets. The outcomes indicate that the adaptability of AV-CANet to different fusion strategies has crucial influence on performance. \textbf{(4)} Afterwards, we present the ablation explorations on the two sub-modules of LGF Module, as depicted in Tab.~\ref{tab-LISF-GLCF}. From the results, we conclude that the joint use of two sub-modules leads to the highest improvement, demonstrating their superiority in correlations modeling of audio-visual cues. \textbf{(5)} Finally, we ablate the emotion polarities with different penalties, as shown in Tab.~\ref{tab-ep-ce-loss}. We find that when $\gamma_{\textit{ep}(y_i)}$ = 0.7, AV-CANet consistently performs best across three datasets. We thus determine that larger penalties generally facilitate model to focus more on misclassified samples, while enhancing its capability in learning emotion-related representations. Besides, we observe that smaller penalties on large-scale dataset may confuse model learning, leading to slight declines.

\begin{table}[]
\centering
\setlength{\abovecaptionskip}{-0em} 
\renewcommand{\arraystretch}{0.90} 
\setlength{\arrayrulewidth}{0.22pt}
\caption{The ablation comparisons of five different fusion strategies in the GLCF Sub-Module. EW: Element-Wise.}
\label{tab-fusion-strategies}
\small
\resizebox{\linewidth}{!}{%
\begin{tabular}{ccccccc}
\toprule[0.22pt]
\multirow{2}{*}{\raisebox{-2ex}{\begin{tabular}[c]{@{}c@{}}Fusion \\Strategy\end{tabular}}} & \multicolumn{2}{c}{eMotions}    & \multicolumn{2}{c}{Music\_video} & \multicolumn{2}{c}{Ekman6}      \\ \cmidrule[0.22pt](l){2-3}   \cmidrule[0.22pt](l){4-5}   \cmidrule[0.22pt](l){6-7} 
                                                                            & ACC         & WA-F1          & ACC         & WA-F1          & ACC         & WA-F1          \\ \midrule[0.22pt]
Gated \cite{60_kiela2018efficient}                                                                & 67.39          & \textbf{67.35}          & 80.51          & 80.26 & 50.94          & 50.45          \\
EW-Multiply          & 64.44          & 63.97          & 76.71          & 76.44          & 48.44          & 47.17          \\
Neural                                                                & 66.52          & 66.12          & 82.03          & 81.75          & 53.44          & 52.34          \\
\rowcolor{gray!20}
Mid-Concat                                                         & \textbf{67.48}          & 67.19          & \textbf{82.78}          & \textbf{82.46}          & \textbf{54.06}          & \textbf{53.51}          \\
Sum                                                                   & 66.49 & 66.17 & 81.77 & 81.64          & 50.94 & 50.18 \\ \bottomrule[0.22pt]
\end{tabular}
}
\vspace{-0.8em}
\end{table}

\begin{table}[]
\centering
\setlength{\abovecaptionskip}{-0em} 
\renewcommand{\arraystretch}{0.90}
\setlength{\arrayrulewidth}{0.22pt}
\caption{Ablation study for the LISF and GLCF Sub-Modules.}
\small
\resizebox{\linewidth}{!}{%
\begin{tabular}{ccccccc}
\toprule[0.22pt]
\multirow{2}{*}{\makecell{LG-CF \\ Module}} & \multicolumn{2}{c}{eMotions} & \multicolumn{2}{c}{Music\_video} & \multicolumn{2}{c}{Ekman6} \\ 
\cmidrule[0.22pt](l){2-3}   \cmidrule[0.22pt](l){4-5}   \cmidrule[0.22pt](l){6-7} 
                                  & ACC        & WA-F1        & ACC          & WA-F1         & ACC       & WA-F1       \\ \midrule[0.22pt]
None                         & 65.98         & 65.05        & 75.19           & 74.66         & 50.31        & 50.41       \\
LISF                        & 67.41         & 67.07        & 82.44           & 81.93         & 53.44        & 52.52       \\
GLCF                      & 67.08         & 66.41        & 81.89           & 81.00         & 51.88        & 51.30       \\
\rowcolor{gray!20}
LISF + GLCF                       & \textbf{67.48}         & \textbf{67.19}        & \textbf{82.78}           & \textbf{82.46}         & \textbf{54.06}        & \textbf{53.51}       \\ \hline
\end{tabular}
}
\label{tab-LISF-GLCF}
\vspace{-1.2em}
\end{table}

\begin{table}[]
\centering
\setlength{\abovecaptionskip}{-0em} 
\renewcommand{\arraystretch}{0.90} 
\setlength{\arrayrulewidth}{0.22pt}
\caption{The ablation comparisons of five different penalties for EP-CE Loss. Note that $\gamma_{neu}$ doesn't function on Ekman6.}
\label{tab-ep-ce-loss}
\small
\resizebox{\linewidth}{!}{%
\begin{tabular}{ccccccc}
\toprule[0.22pt]
\multirow{2}{*}{$\gamma_{pos}$: $\gamma_{neu}$: $\gamma_{neg}$} & \multicolumn{2}{c}{eMotions} & \multicolumn{2}{c}{Music\_video} & \multicolumn{2}{c}{Ekman6} \\ 
\cmidrule[0.22pt](l){2-3}   \cmidrule[0.22pt](l){4-5}   \cmidrule[0.22pt](l){6-7} 
                                  & ACC        & WA-F1        & ACC          & WA-F1         & ACC       & WA-F1       \\ \midrule[0.22pt]
0.3 : \textcolor{gray}{0.3} : 0.3                         & 67.34         & 66.83        & 84.03           & 83.75         & 54.37        & 54.18       \\
0.4 : \textcolor{gray}{0.4} : 0.4                        & 67.34         & 66.89        & 84.05           & 83.89         & 54.38        & 53.88       \\
0.5 : \textcolor{gray}{0.5} : 0.5                       & 67.53         & 67.22        & 84.81           & 84.58         & 55.00        & 54.29       \\
\rowcolor{gray!20}
0.7 : \textcolor{gray}{0.7} : 0.7                       & \textbf{67.79}         & \textbf{67.32}        & \textbf{84.81}           & \textbf{84.63}         & \textbf{55.63}        & \textbf{55.44}       \\ \hline
\end{tabular}
}
\vspace{-1.3em}
\end{table}

\section{Conclusion and Prospects}
\label{sec:discussion}
In this paper, we introduce eMotions, the first large-scale VEA dataset towards SVs. It comprises 27,996 videos labeled across six emotions, sourcing from three SVs platforms. Meanwhile, we make efforts to augment the labeling quality by alleviating the influence of subjectivities. Additionally, two variant datasets are provided through targeted data sampling. We also develop the baseline AV-CANet to tackle the inherent challenges of VEA towards SVs. Extensive experimental results on three eMotions-related and four public datasets verify the superiority of our model. We hope this work can serve as a foundation and inspire more research.

The granular divisions of audio facilitate models to output more refined representations, and the utilization of text overlays in SVs can promote the understanding of emotions, both indicating the potential trends of eMotions. Moreover, we focus on improving the limited cultural generalizability and long-tail distributions presented in eMotions for future developments.

\begin{acks}
eMotions and its two variants will not be transferred to outside parties without permissions and can be only utilized for academic research. In particular, the datasets will not be included as a part of any commercial software package or product of any institution.
\end{acks}

\bibliographystyle{ACM-Reference-Format}
\bibliography{reference}


\begin{thebibliography}{68}


\ifx \showCODEN    \undefined \def \showCODEN     #1{\unskip}     \fi
\ifx \showDOI      \undefined \def \showDOI       #1{#1}\fi
\ifx \showISBNx    \undefined \def \showISBNx     #1{\unskip}     \fi
\ifx \showISBNxiii \undefined \def \showISBNxiii  #1{\unskip}     \fi
\ifx \showISSN     \undefined \def \showISSN      #1{\unskip}     \fi
\ifx \showLCCN     \undefined \def \showLCCN      #1{\unskip}     \fi
\ifx \shownote     \undefined \def \shownote      #1{#1}          \fi
\ifx \showarticletitle \undefined \def \showarticletitle #1{#1}   \fi
\ifx \showURL      \undefined \def \showURL       {\relax}        \fi
\providecommand\bibfield[2]{#2}
\providecommand\bibinfo[2]{#2}
\providecommand\natexlab[1]{#1}
\providecommand\showeprint[2][]{arXiv:#2}

\bibitem[Abercrombie et~al\mbox{.}(2023)]%
        {new_28_abercrombie2023consistency}
\bibfield{author}{\bibinfo{person}{Gavin Abercrombie}, \bibinfo{person}{Verena Rieser}, {and} \bibinfo{person}{Dirk Hovy}.} \bibinfo{year}{2023}\natexlab{}.
\newblock \showarticletitle{Consistency is Key: Disentangling Label Variation in Natural Language Processing with Intra-Annotator Agreement}.
\newblock \bibinfo{journal}{\emph{arXiv preprint arXiv:2301.10684}} (\bibinfo{year}{2023}).
\newblock


\bibitem[Ba et~al\mbox{.}(2016)]%
        {new_50_ba2016layer}
\bibfield{author}{\bibinfo{person}{Jimmy~Lei Ba}, \bibinfo{person}{Jamie~Ryan Kiros}, {and} \bibinfo{person}{Geoffrey~E Hinton}.} \bibinfo{year}{2016}\natexlab{}.
\newblock \showarticletitle{Layer normalization}.
\newblock \bibinfo{journal}{\emph{arXiv preprint arXiv:1607.06450}} (\bibinfo{year}{2016}).
\newblock


\bibitem[Bertasius et~al\mbox{.}(2021)]%
        {30_bertasius2021space}
\bibfield{author}{\bibinfo{person}{Gedas Bertasius}, \bibinfo{person}{Heng Wang}, {and} \bibinfo{person}{Lorenzo Torresani}.} \bibinfo{year}{2021}\natexlab{}.
\newblock \showarticletitle{Is space-time attention all you need for video understanding?}. In \bibinfo{booktitle}{\emph{ICML}}, Vol.~\bibinfo{volume}{2}. \bibinfo{pages}{4}.
\newblock


\bibitem[Busso et~al\mbox{.}(2008)]%
        {26_busso2008iemocap}
\bibfield{author}{\bibinfo{person}{Carlos Busso}, \bibinfo{person}{Murtaza Bulut}, \bibinfo{person}{Chi-Chun Lee}, \bibinfo{person}{Abe Kazemzadeh}, \bibinfo{person}{Emily Mower}, \bibinfo{person}{Samuel Kim}, \bibinfo{person}{Jeannette~N Chang}, \bibinfo{person}{Sungbok Lee}, {and} \bibinfo{person}{Shrikanth~S Narayanan}.} \bibinfo{year}{2008}\natexlab{}.
\newblock \showarticletitle{IEMOCAP: Interactive emotional dyadic motion capture database}.
\newblock \bibinfo{journal}{\emph{Language resources and evaluation}}  \bibinfo{volume}{42} (\bibinfo{year}{2008}), \bibinfo{pages}{335--359}.
\newblock


\bibitem[Carreira and Zisserman(2017)]%
        {43_carreira2017quo}
\bibfield{author}{\bibinfo{person}{Joao Carreira} {and} \bibinfo{person}{Andrew Zisserman}.} \bibinfo{year}{2017}\natexlab{}.
\newblock \showarticletitle{Quo vadis, action recognition? a new model and the kinetics dataset}. In \bibinfo{booktitle}{\emph{proceedings of the IEEE Conference on Computer Vision and Pattern Recognition}}. \bibinfo{pages}{6299--6308}.
\newblock


\bibitem[Chumachenko et~al\mbox{.}(2022)]%
        {new_18_chumachenko2022self}
\bibfield{author}{\bibinfo{person}{Kateryna Chumachenko}, \bibinfo{person}{Alexandros Iosifidis}, {and} \bibinfo{person}{Moncef Gabbouj}.} \bibinfo{year}{2022}\natexlab{}.
\newblock \showarticletitle{Self-attention fusion for audiovisual emotion recognition with incomplete data}. In \bibinfo{booktitle}{\emph{2022 26th International Conference on Pattern Recognition (ICPR)}}. IEEE, \bibinfo{pages}{2822--2828}.
\newblock


\bibitem[Desplanques et~al\mbox{.}(2020)]%
        {new_32_desplanques2020ecapa}
\bibfield{author}{\bibinfo{person}{Brecht Desplanques}, \bibinfo{person}{Jenthe Thienpondt}, {and} \bibinfo{person}{Kris Demuynck}.} \bibinfo{year}{2020}\natexlab{}.
\newblock \showarticletitle{Ecapa-tdnn: Emphasized channel attention, propagation and aggregation in tdnn based speaker verification}.
\newblock \bibinfo{journal}{\emph{arXiv preprint arXiv:2005.07143}} (\bibinfo{year}{2020}).
\newblock


\bibitem[Dhall et~al\mbox{.}(2012)]%
        {new_6_2012Collecting}
\bibfield{author}{\bibinfo{person}{Abhinav Dhall}, \bibinfo{person}{Roland Goecke}, \bibinfo{person}{Simon Lucey}, {and} \bibinfo{person}{Tom Gedeon}.} \bibinfo{year}{2012}\natexlab{}.
\newblock \showarticletitle{Collecting Large, Richly Annotated Facial-Expression Databases from Movies}.
\newblock \bibinfo{journal}{\emph{IEEE Multimedia}} \bibinfo{volume}{19}, \bibinfo{number}{3} (\bibinfo{year}{2012}), \bibinfo{pages}{0034}.
\newblock


\bibitem[Feichtenhofer et~al\mbox{.}(2019)]%
        {new_33_feichtenhofer2019slowfast}
\bibfield{author}{\bibinfo{person}{Christoph Feichtenhofer}, \bibinfo{person}{Haoqi Fan}, \bibinfo{person}{Jitendra Malik}, {and} \bibinfo{person}{Kaiming He}.} \bibinfo{year}{2019}\natexlab{}.
\newblock \showarticletitle{Slowfast networks for video recognition}. In \bibinfo{booktitle}{\emph{Proceedings of the IEEE/CVF international conference on computer vision}}. \bibinfo{pages}{6202--6211}.
\newblock


\bibitem[Frisch and Giulianelli(2024)]%
        {new_39_frisch2024llm}
\bibfield{author}{\bibinfo{person}{Ivar Frisch} {and} \bibinfo{person}{Mario Giulianelli}.} \bibinfo{year}{2024}\natexlab{}.
\newblock \showarticletitle{LLM Agents in Interaction: Measuring Personality Consistency and Linguistic Alignment in Interacting Populations of Large Language Models}.
\newblock \bibinfo{journal}{\emph{arXiv preprint arXiv:2402.02896}} (\bibinfo{year}{2024}).
\newblock


\bibitem[Gao et~al\mbox{.}(2019)]%
        {34_gao2019res2net}
\bibfield{author}{\bibinfo{person}{Shang-Hua Gao}, \bibinfo{person}{Ming-Ming Cheng}, \bibinfo{person}{Kai Zhao}, \bibinfo{person}{Xin-Yu Zhang}, \bibinfo{person}{Ming-Hsuan Yang}, {and} \bibinfo{person}{Philip Torr}.} \bibinfo{year}{2019}\natexlab{}.
\newblock \showarticletitle{Res2net: A new multi-scale backbone architecture}.
\newblock \bibinfo{journal}{\emph{IEEE transactions on pattern analysis and machine intelligence}} \bibinfo{volume}{43}, \bibinfo{number}{2} (\bibinfo{year}{2019}), \bibinfo{pages}{652--662}.
\newblock


\bibitem[Gunes and Piccardi(2006)]%
        {new_9_gunes2006bimodal}
\bibfield{author}{\bibinfo{person}{Hatice Gunes} {and} \bibinfo{person}{Massimo Piccardi}.} \bibinfo{year}{2006}\natexlab{}.
\newblock \showarticletitle{A bimodal face and body gesture database for automatic analysis of human nonverbal affective behavior}. In \bibinfo{booktitle}{\emph{18th International conference on pattern recognition (ICPR'06)}}, Vol.~\bibinfo{volume}{1}. IEEE, \bibinfo{pages}{1148--1153}.
\newblock


\bibitem[Hallgren(2012)]%
        {new_27_hallgren2012computing}
\bibfield{author}{\bibinfo{person}{Kevin~A Hallgren}.} \bibinfo{year}{2012}\natexlab{}.
\newblock \showarticletitle{Computing inter-rater reliability for observational data: an overview and tutorial}.
\newblock \bibinfo{journal}{\emph{Tutorials in quantitative methods for psychology}} \bibinfo{volume}{8}, \bibinfo{number}{1} (\bibinfo{year}{2012}), \bibinfo{pages}{23}.
\newblock


\bibitem[Hanjalic and Xu(2005)]%
        {new_51_hanjalic2005affective}
\bibfield{author}{\bibinfo{person}{Alan Hanjalic} {and} \bibinfo{person}{Li-Qun Xu}.} \bibinfo{year}{2005}\natexlab{}.
\newblock \showarticletitle{Affective video content representation and modeling}.
\newblock \bibinfo{journal}{\emph{IEEE transactions on multimedia}} \bibinfo{volume}{7}, \bibinfo{number}{1} (\bibinfo{year}{2005}), \bibinfo{pages}{143--154}.
\newblock


\bibitem[Hara et~al\mbox{.}(2018)]%
        {21_hara2018can}
\bibfield{author}{\bibinfo{person}{Kensho Hara}, \bibinfo{person}{Hirokatsu Kataoka}, {and} \bibinfo{person}{Yutaka Satoh}.} \bibinfo{year}{2018}\natexlab{}.
\newblock \showarticletitle{Can spatiotemporal 3d cnns retrace the history of 2d cnns and imagenet?}. In \bibinfo{booktitle}{\emph{Proceedings of the IEEE conference on Computer Vision and Pattern Recognition}}. \bibinfo{pages}{6546--6555}.
\newblock


\bibitem[He et~al\mbox{.}(2016)]%
        {22_he2016deep}
\bibfield{author}{\bibinfo{person}{Kaiming He}, \bibinfo{person}{Xiangyu Zhang}, \bibinfo{person}{Shaoqing Ren}, {and} \bibinfo{person}{Jian Sun}.} \bibinfo{year}{2016}\natexlab{}.
\newblock \showarticletitle{Deep residual learning for image recognition}. In \bibinfo{booktitle}{\emph{Proceedings of the IEEE conference on computer vision and pattern recognition}}. \bibinfo{pages}{770--778}.
\newblock


\bibitem[Jiang et~al\mbox{.}(2020)]%
        {new_58_jiang2020dfew}
\bibfield{author}{\bibinfo{person}{Xingxun Jiang}, \bibinfo{person}{Yuan Zong}, \bibinfo{person}{Wenming Zheng}, \bibinfo{person}{Chuangao Tang}, \bibinfo{person}{Wanchuang Xia}, \bibinfo{person}{Cheng Lu}, {and} \bibinfo{person}{Jiateng Liu}.} \bibinfo{year}{2020}\natexlab{}.
\newblock \showarticletitle{Dfew: A large-scale database for recognizing dynamic facial expressions in the wild}. In \bibinfo{booktitle}{\emph{Proceedings of the 28th ACM international conference on multimedia}}. \bibinfo{pages}{2881--2889}.
\newblock


\bibitem[Jiang et~al\mbox{.}(2014)]%
        {12_jiang2014predicting}
\bibfield{author}{\bibinfo{person}{Yu-Gang Jiang}, \bibinfo{person}{Baohan Xu}, {and} \bibinfo{person}{Xiangyang Xue}.} \bibinfo{year}{2014}\natexlab{}.
\newblock \showarticletitle{Predicting emotions in user-generated videos}. In \bibinfo{booktitle}{\emph{Proceedings of the AAAI conference on artificial intelligence}}, Vol.~\bibinfo{volume}{28}.
\newblock


\bibitem[Kiela et~al\mbox{.}(2018)]%
        {60_kiela2018efficient}
\bibfield{author}{\bibinfo{person}{Douwe Kiela}, \bibinfo{person}{Edouard Grave}, \bibinfo{person}{Armand Joulin}, {and} \bibinfo{person}{Tomas Mikolov}.} \bibinfo{year}{2018}\natexlab{}.
\newblock \showarticletitle{Efficient large-scale multi-modal classification}. In \bibinfo{booktitle}{\emph{Proceedings of the AAAI conference on artificial intelligence}}, Vol.~\bibinfo{volume}{32}.
\newblock


\bibitem[Kollias and Zafeiriou(2018)]%
        {new_8_kollias2018aff}
\bibfield{author}{\bibinfo{person}{Dimitrios Kollias} {and} \bibinfo{person}{Stefanos Zafeiriou}.} \bibinfo{year}{2018}\natexlab{}.
\newblock \showarticletitle{Aff-wild2: Extending the aff-wild database for affect recognition}.
\newblock \bibinfo{journal}{\emph{arXiv preprint arXiv:1811.07770}} (\bibinfo{year}{2018}).
\newblock


\bibitem[Kong et~al\mbox{.}(2020)]%
        {27_kong2020panns}
\bibfield{author}{\bibinfo{person}{Qiuqiang Kong}, \bibinfo{person}{Yin Cao}, \bibinfo{person}{Turab Iqbal}, \bibinfo{person}{Yuxuan Wang}, \bibinfo{person}{Wenwu Wang}, {and} \bibinfo{person}{Mark~D Plumbley}.} \bibinfo{year}{2020}\natexlab{}.
\newblock \showarticletitle{Panns: Large-scale pretrained audio neural networks for audio pattern recognition}.
\newblock \bibinfo{journal}{\emph{IEEE/ACM Transactions on Audio, Speech, and Language Processing}}  \bibinfo{volume}{28} (\bibinfo{year}{2020}), \bibinfo{pages}{2880--2894}.
\newblock


\bibitem[Lavitas et~al\mbox{.}(2021)]%
        {new_29_lavitas2021annotation}
\bibfield{author}{\bibinfo{person}{Liliya Lavitas}, \bibinfo{person}{Olivia Redfield}, \bibinfo{person}{Allen Lee}, \bibinfo{person}{Daniel Fletcher}, \bibinfo{person}{Matthias Eck}, {and} \bibinfo{person}{Sunil Janardhanan}.} \bibinfo{year}{2021}\natexlab{}.
\newblock \showarticletitle{Annotation quality framework-accuracy, credibility, and consistency}. In \bibinfo{booktitle}{\emph{NEURIPS 2021 Workshop for Data Centric AI}}.
\newblock


\bibitem[Lee et~al\mbox{.}(2019)]%
        {18_lee2019context}
\bibfield{author}{\bibinfo{person}{Jiyoung Lee}, \bibinfo{person}{Seungryong Kim}, \bibinfo{person}{Sunok Kim}, \bibinfo{person}{Jungin Park}, {and} \bibinfo{person}{Kwanghoon Sohn}.} \bibinfo{year}{2019}\natexlab{}.
\newblock \showarticletitle{Context-aware emotion recognition networks}. In \bibinfo{booktitle}{\emph{Proceedings of the IEEE/CVF international conference on computer vision}}. \bibinfo{pages}{10143--10152}.
\newblock


\bibitem[Li et~al\mbox{.}(2020)]%
        {new_38_li2020ms}
\bibfield{author}{\bibinfo{person}{Shijie Li}, \bibinfo{person}{Yazan~Abu Farha}, \bibinfo{person}{Yun Liu}, \bibinfo{person}{Ming-Ming Cheng}, {and} \bibinfo{person}{Juergen Gall}.} \bibinfo{year}{2020}\natexlab{}.
\newblock \showarticletitle{Ms-tcn++: Multi-stage temporal convolutional network for action segmentation}.
\newblock \bibinfo{journal}{\emph{IEEE transactions on pattern analysis and machine intelligence}} \bibinfo{volume}{45}, \bibinfo{number}{6} (\bibinfo{year}{2020}), \bibinfo{pages}{6647--6658}.
\newblock


\bibitem[Lian et~al\mbox{.}(2024)]%
        {new_42_lian2024mer}
\bibfield{author}{\bibinfo{person}{Zheng Lian}, \bibinfo{person}{Haiyang Sun}, \bibinfo{person}{Licai Sun}, \bibinfo{person}{Zhuofan Wen}, \bibinfo{person}{Siyuan Zhang}, \bibinfo{person}{Shun Chen}, \bibinfo{person}{Hao Gu}, \bibinfo{person}{Jinming Zhao}, \bibinfo{person}{Ziyang Ma}, \bibinfo{person}{Xie Chen}, {et~al\mbox{.}}} \bibinfo{year}{2024}\natexlab{}.
\newblock \showarticletitle{MER 2024: Semi-Supervised Learning, Noise Robustness, and Open-Vocabulary Multimodal Emotion Recognition}.
\newblock \bibinfo{journal}{\emph{arXiv preprint arXiv:2404.17113}} (\bibinfo{year}{2024}).
\newblock


\bibitem[Lian et~al\mbox{.}(2023)]%
        {new_56_lian2023explainable}
\bibfield{author}{\bibinfo{person}{Zheng Lian}, \bibinfo{person}{Licai Sun}, \bibinfo{person}{Mingyu Xu}, \bibinfo{person}{Haiyang Sun}, \bibinfo{person}{Ke Xu}, \bibinfo{person}{Zhuofan Wen}, \bibinfo{person}{Shun Chen}, \bibinfo{person}{Bin Liu}, {and} \bibinfo{person}{Jianhua Tao}.} \bibinfo{year}{2023}\natexlab{}.
\newblock \showarticletitle{Explainable multimodal emotion reasoning}.
\newblock \bibinfo{journal}{\emph{arXiv preprint arXiv:2306.15401}} (\bibinfo{year}{2023}).
\newblock


\bibitem[Liu et~al\mbox{.}(2022d)]%
        {new_21_liu2022ser30k}
\bibfield{author}{\bibinfo{person}{Shengzhe Liu}, \bibinfo{person}{Xin Zhang}, {and} \bibinfo{person}{Jufeng Yang}.} \bibinfo{year}{2022}\natexlab{d}.
\newblock \showarticletitle{SER30K: A large-scale dataset for sticker emotion recognition}. In \bibinfo{booktitle}{\emph{Proceedings of the 30th ACM International Conference on Multimedia}}. \bibinfo{pages}{33--41}.
\newblock


\bibitem[Liu et~al\mbox{.}(2022a)]%
        {new_57_liu2022mafw}
\bibfield{author}{\bibinfo{person}{Yuanyuan Liu}, \bibinfo{person}{Wei Dai}, \bibinfo{person}{Chuanxu Feng}, \bibinfo{person}{Wenbin Wang}, \bibinfo{person}{Guanghao Yin}, \bibinfo{person}{Jiabei Zeng}, {and} \bibinfo{person}{Shiguang Shan}.} \bibinfo{year}{2022}\natexlab{a}.
\newblock \showarticletitle{Mafw: A large-scale, multi-modal, compound affective database for dynamic facial expression recognition in the wild}. In \bibinfo{booktitle}{\emph{Proceedings of the 30th ACM International Conference on Multimedia}}. \bibinfo{pages}{24--32}.
\newblock


\bibitem[Liu et~al\mbox{.}(2022b)]%
        {46_liu2022convnet}
\bibfield{author}{\bibinfo{person}{Zhuang Liu}, \bibinfo{person}{Hanzi Mao}, \bibinfo{person}{Chao-Yuan Wu}, \bibinfo{person}{Christoph Feichtenhofer}, \bibinfo{person}{Trevor Darrell}, {and} \bibinfo{person}{Saining Xie}.} \bibinfo{year}{2022}\natexlab{b}.
\newblock \showarticletitle{A convnet for the 2020s}. In \bibinfo{booktitle}{\emph{Proceedings of the IEEE/CVF conference on computer vision and pattern recognition}}. \bibinfo{pages}{11976--11986}.
\newblock


\bibitem[Liu et~al\mbox{.}(2022c)]%
        {19_liu2022video}
\bibfield{author}{\bibinfo{person}{Ze Liu}, \bibinfo{person}{Jia Ning}, \bibinfo{person}{Yue Cao}, \bibinfo{person}{Yixuan Wei}, \bibinfo{person}{Zheng Zhang}, \bibinfo{person}{Stephen Lin}, {and} \bibinfo{person}{Han Hu}.} \bibinfo{year}{2022}\natexlab{c}.
\newblock \showarticletitle{Video swin transformer}. In \bibinfo{booktitle}{\emph{Proceedings of the IEEE/CVF conference on computer vision and pattern recognition}}. \bibinfo{pages}{3202--3211}.
\newblock


\bibitem[Mayer et~al\mbox{.}(2002)]%
        {2_mayer2002mayer}
\bibfield{author}{\bibinfo{person}{John~D Mayer}, \bibinfo{person}{Peter Salovey}, {and} \bibinfo{person}{David~R Caruso}.} \bibinfo{year}{2002}\natexlab{}.
\newblock \showarticletitle{Mayer-Salovey-Caruso emotional intelligence test (MSCEIT) users manual}.
\newblock  (\bibinfo{year}{2002}).
\newblock


\bibitem[Mocanu et~al\mbox{.}(2023)]%
        {new_13_mocanu2023multimodal}
\bibfield{author}{\bibinfo{person}{Bogdan Mocanu}, \bibinfo{person}{Ruxandra Tapu}, {and} \bibinfo{person}{Titus Zaharia}.} \bibinfo{year}{2023}\natexlab{}.
\newblock \showarticletitle{Multimodal emotion recognition using cross modal audio-video fusion with attention and deep metric learning}.
\newblock \bibinfo{journal}{\emph{Image and Vision Computing}}  \bibinfo{volume}{133} (\bibinfo{year}{2023}), \bibinfo{pages}{104676}.
\newblock


\bibitem[Pandeya and Lee(2021)]%
        {7_pandeya2021deep}
\bibfield{author}{\bibinfo{person}{Yagya~Raj Pandeya} {and} \bibinfo{person}{Joonwhoan Lee}.} \bibinfo{year}{2021}\natexlab{}.
\newblock \showarticletitle{Deep learning-based late fusion of multimodal information for emotion classification of music video}.
\newblock \bibinfo{journal}{\emph{Multimedia Tools and Applications}}  \bibinfo{volume}{80} (\bibinfo{year}{2021}), \bibinfo{pages}{2887--2905}.
\newblock


\bibitem[Peng et~al\mbox{.}(2022)]%
        {50_peng2022balanced}
\bibfield{author}{\bibinfo{person}{Xiaokang Peng}, \bibinfo{person}{Yake Wei}, \bibinfo{person}{Andong Deng}, \bibinfo{person}{Dong Wang}, {and} \bibinfo{person}{Di Hu}.} \bibinfo{year}{2022}\natexlab{}.
\newblock \showarticletitle{Balanced multimodal learning via on-the-fly gradient modulation}. In \bibinfo{booktitle}{\emph{Proceedings of the IEEE/CVF Conference on Computer Vision and Pattern Recognition}}. \bibinfo{pages}{8238--8247}.
\newblock


\bibitem[Plutchik(1994)]%
        {1_plutchik1994psychology}
\bibfield{author}{\bibinfo{person}{Robert Plutchik}.} \bibinfo{year}{1994}\natexlab{}.
\newblock \bibinfo{booktitle}{\emph{The psychology and biology of emotion.}}
\newblock \bibinfo{publisher}{HarperCollins College Publishers}.
\newblock


\bibitem[Poria et~al\mbox{.}(2017)]%
        {55_poria2017context}
\bibfield{author}{\bibinfo{person}{Soujanya Poria}, \bibinfo{person}{Erik Cambria}, \bibinfo{person}{Devamanyu Hazarika}, \bibinfo{person}{Navonil Majumder}, \bibinfo{person}{Amir Zadeh}, {and} \bibinfo{person}{Louis-Philippe Morency}.} \bibinfo{year}{2017}\natexlab{}.
\newblock \showarticletitle{Context-dependent sentiment analysis in user-generated videos}. In \bibinfo{booktitle}{\emph{Proceedings of the 55th annual meeting of the association for computational linguistics (volume 1: Long papers)}}. \bibinfo{pages}{873--883}.
\newblock


\bibitem[Poria et~al\mbox{.}(2018)]%
        {33_soujanya2018multimodal}
\bibfield{author}{\bibinfo{person}{Soujanya Poria}, \bibinfo{person}{Devamanyu Hazarika}, \bibinfo{person}{Navonil Majumder}, \bibinfo{person}{Gautam Naik}, \bibinfo{person}{Erik Cambria}, {and} \bibinfo{person}{Rada Mihalcea}.} \bibinfo{year}{2018}\natexlab{}.
\newblock \showarticletitle{Meld: A multimodal multi-party dataset for emotion recognition in conversations}.
\newblock \bibinfo{journal}{\emph{arXiv preprint arXiv:1810.02508}} (\bibinfo{year}{2018}).
\newblock


\bibitem[Posner et~al\mbox{.}(2005)]%
        {new_60_posner2005circumplex}
\bibfield{author}{\bibinfo{person}{Jonathan Posner}, \bibinfo{person}{James~A Russell}, {and} \bibinfo{person}{Bradley~S Peterson}.} \bibinfo{year}{2005}\natexlab{}.
\newblock \showarticletitle{The circumplex model of affect: An integrative approach to affective neuroscience, cognitive development, and psychopathology}.
\newblock \bibinfo{journal}{\emph{Development and psychopathology}} \bibinfo{volume}{17}, \bibinfo{number}{3} (\bibinfo{year}{2005}), \bibinfo{pages}{715--734}.
\newblock


\bibitem[Praveen et~al\mbox{.}(2023)]%
        {new_14_praveen2023audio}
\bibfield{author}{\bibinfo{person}{R~Gnana Praveen}, \bibinfo{person}{Patrick Cardinal}, {and} \bibinfo{person}{Eric Granger}.} \bibinfo{year}{2023}\natexlab{}.
\newblock \showarticletitle{Audio-visual fusion for emotion recognition in the valence-arousal space using joint cross-attention}.
\newblock \bibinfo{journal}{\emph{IEEE Transactions on Biometrics, Behavior, and Identity Science}} (\bibinfo{year}{2023}).
\newblock


\bibitem[Radford et~al\mbox{.}(2021)]%
        {new_43_radford2021learning}
\bibfield{author}{\bibinfo{person}{Alec Radford}, \bibinfo{person}{Jong~Wook Kim}, \bibinfo{person}{Chris Hallacy}, \bibinfo{person}{Aditya Ramesh}, \bibinfo{person}{Gabriel Goh}, \bibinfo{person}{Sandhini Agarwal}, \bibinfo{person}{Girish Sastry}, \bibinfo{person}{Amanda Askell}, \bibinfo{person}{Pamela Mishkin}, \bibinfo{person}{Jack Clark}, {et~al\mbox{.}}} \bibinfo{year}{2021}\natexlab{}.
\newblock \showarticletitle{Learning transferable visual models from natural language supervision}. In \bibinfo{booktitle}{\emph{International conference on machine learning}}. PMLR, \bibinfo{pages}{8748--8763}.
\newblock


\bibitem[Ribeiro et~al\mbox{.}(2011)]%
        {new_24_ribeiro2011crowdsourcing}
\bibfield{author}{\bibinfo{person}{Fl{\'a}vio Ribeiro}, \bibinfo{person}{Dinei Florencio}, {and} \bibinfo{person}{V{\'\i}tor Nascimento}.} \bibinfo{year}{2011}\natexlab{}.
\newblock \showarticletitle{Crowdsourcing subjective image quality evaluation}. In \bibinfo{booktitle}{\emph{2011 18th IEEE International Conference on Image Processing}}. IEEE, \bibinfo{pages}{3097--3100}.
\newblock


\bibitem[Schoneveld et~al\mbox{.}(2021)]%
        {new_17_schoneveld2021leveraging}
\bibfield{author}{\bibinfo{person}{Liam Schoneveld}, \bibinfo{person}{Alice Othmani}, {and} \bibinfo{person}{Hazem Abdelkawy}.} \bibinfo{year}{2021}\natexlab{}.
\newblock \showarticletitle{Leveraging recent advances in deep learning for audio-visual emotion recognition}.
\newblock \bibinfo{journal}{\emph{Pattern Recognition Letters}}  \bibinfo{volume}{146} (\bibinfo{year}{2021}), \bibinfo{pages}{1--7}.
\newblock


\bibitem[Schuller and Schuller(2018)]%
        {5_schuller2018age}
\bibfield{author}{\bibinfo{person}{Dagmar Schuller} {and} \bibinfo{person}{Bj{\"o}rn~W Schuller}.} \bibinfo{year}{2018}\natexlab{}.
\newblock \showarticletitle{The age of artificial emotional intelligence}.
\newblock \bibinfo{journal}{\emph{Computer}} \bibinfo{volume}{51}, \bibinfo{number}{9} (\bibinfo{year}{2018}), \bibinfo{pages}{38--46}.
\newblock


\bibitem[Singh and Masuku(2014)]%
        {new_23_singh2014sampling}
\bibfield{author}{\bibinfo{person}{Ajay~S Singh} {and} \bibinfo{person}{Micah~B Masuku}.} \bibinfo{year}{2014}\natexlab{}.
\newblock \showarticletitle{Sampling techniques \& determination of sample size in applied statistics research: An overview}.
\newblock \bibinfo{journal}{\emph{International Journal of economics, commerce and management}} \bibinfo{volume}{2}, \bibinfo{number}{11} (\bibinfo{year}{2014}), \bibinfo{pages}{1--22}.
\newblock


\bibitem[Sobkowicz et~al\mbox{.}(2012)]%
        {new_54_sobkowicz2012opinion}
\bibfield{author}{\bibinfo{person}{Pawel Sobkowicz}, \bibinfo{person}{Michael Kaschesky}, {and} \bibinfo{person}{Guillaume Bouchard}.} \bibinfo{year}{2012}\natexlab{}.
\newblock \showarticletitle{Opinion mining in social media: Modeling, simulating, and forecasting political opinions in the web}.
\newblock \bibinfo{journal}{\emph{Government information quarterly}} \bibinfo{volume}{29}, \bibinfo{number}{4} (\bibinfo{year}{2012}), \bibinfo{pages}{470--479}.
\newblock


\bibitem[Su et~al\mbox{.}(2020)]%
        {new_35_su2020msaf}
\bibfield{author}{\bibinfo{person}{Lang Su}, \bibinfo{person}{Chuqing Hu}, \bibinfo{person}{Guofa Li}, {and} \bibinfo{person}{Dongpu Cao}.} \bibinfo{year}{2020}\natexlab{}.
\newblock \showarticletitle{Msaf: Multimodal split attention fusion}.
\newblock \bibinfo{journal}{\emph{arXiv preprint arXiv:2012.07175}} (\bibinfo{year}{2020}).
\newblock


\bibitem[Sun et~al\mbox{.}(2023)]%
        {new_52_sun2023mae}
\bibfield{author}{\bibinfo{person}{Licai Sun}, \bibinfo{person}{Zheng Lian}, \bibinfo{person}{Bin Liu}, {and} \bibinfo{person}{Jianhua Tao}.} \bibinfo{year}{2023}\natexlab{}.
\newblock \showarticletitle{Mae-dfer: Efficient masked autoencoder for self-supervised dynamic facial expression recognition}. In \bibinfo{booktitle}{\emph{Proceedings of the 31st ACM International Conference on Multimedia}}. \bibinfo{pages}{6110--6121}.
\newblock


\bibitem[Sun et~al\mbox{.}(2024)]%
        {new_45_sun2024hicmae}
\bibfield{author}{\bibinfo{person}{Licai Sun}, \bibinfo{person}{Zheng Lian}, \bibinfo{person}{Bin Liu}, {and} \bibinfo{person}{Jianhua Tao}.} \bibinfo{year}{2024}\natexlab{}.
\newblock \showarticletitle{HiCMAE: Hierarchical Contrastive Masked Autoencoder for Self-Supervised Audio-Visual Emotion Recognition}.
\newblock \bibinfo{journal}{\emph{Information Fusion}}  \bibinfo{volume}{108} (\bibinfo{year}{2024}), \bibinfo{pages}{102382}.
\newblock


\bibitem[Tran et~al\mbox{.}(2015)]%
        {32_tran2015learning}
\bibfield{author}{\bibinfo{person}{Du Tran}, \bibinfo{person}{Lubomir Bourdev}, \bibinfo{person}{Rob Fergus}, \bibinfo{person}{Lorenzo Torresani}, {and} \bibinfo{person}{Manohar Paluri}.} \bibinfo{year}{2015}\natexlab{}.
\newblock \showarticletitle{Learning spatiotemporal features with 3d convolutional networks}. In \bibinfo{booktitle}{\emph{Proceedings of the IEEE international conference on computer vision}}. \bibinfo{pages}{4489--4497}.
\newblock


\bibitem[Tran and Soleymani(2022)]%
        {new_16_tran2022pre}
\bibfield{author}{\bibinfo{person}{Minh Tran} {and} \bibinfo{person}{Mohammad Soleymani}.} \bibinfo{year}{2022}\natexlab{}.
\newblock \showarticletitle{A pre-trained audio-visual transformer for emotion recognition}. In \bibinfo{booktitle}{\emph{ICASSP 2022-2022 IEEE International Conference on Acoustics, Speech and Signal Processing (ICASSP)}}. IEEE, \bibinfo{pages}{4698--4702}.
\newblock


\bibitem[Vaswani et~al\mbox{.}(2017)]%
        {57_vaswani2017attention}
\bibfield{author}{\bibinfo{person}{Ashish Vaswani}, \bibinfo{person}{Noam Shazeer}, \bibinfo{person}{Niki Parmar}, \bibinfo{person}{Jakob Uszkoreit}, \bibinfo{person}{Llion Jones}, \bibinfo{person}{Aidan~N Gomez}, \bibinfo{person}{{\L}ukasz Kaiser}, {and} \bibinfo{person}{Illia Polosukhin}.} \bibinfo{year}{2017}\natexlab{}.
\newblock \showarticletitle{Attention is all you need}.
\newblock \bibinfo{journal}{\emph{Advances in neural information processing systems}}  \bibinfo{volume}{30} (\bibinfo{year}{2017}).
\newblock


\bibitem[Wang et~al\mbox{.}(2020a)]%
        {42_wang2020cspnet}
\bibfield{author}{\bibinfo{person}{Chien-Yao Wang}, \bibinfo{person}{Hong-Yuan~Mark Liao}, \bibinfo{person}{Yueh-Hua Wu}, \bibinfo{person}{Ping-Yang Chen}, \bibinfo{person}{Jun-Wei Hsieh}, {and} \bibinfo{person}{I-Hau Yeh}.} \bibinfo{year}{2020}\natexlab{a}.
\newblock \showarticletitle{CSPNet: A new backbone that can enhance learning capability of CNN}. In \bibinfo{booktitle}{\emph{Proceedings of the IEEE/CVF conference on computer vision and pattern recognition workshops}}. \bibinfo{pages}{390--391}.
\newblock


\bibitem[Wang et~al\mbox{.}(2023)]%
        {38_wang2023cam++}
\bibfield{author}{\bibinfo{person}{Hui Wang}, \bibinfo{person}{Siqi Zheng}, \bibinfo{person}{Yafeng Chen}, \bibinfo{person}{Luyao Cheng}, {and} \bibinfo{person}{Qian Chen}.} \bibinfo{year}{2023}\natexlab{}.
\newblock \showarticletitle{CAM++: A Fast and Efficient Network For Speaker Verification Using Context-Aware Masking}.
\newblock \bibinfo{journal}{\emph{arXiv preprint arXiv:2303.00332}} (\bibinfo{year}{2023}).
\newblock


\bibitem[Wang et~al\mbox{.}(2020b)]%
        {new_1_wang2020suppressing}
\bibfield{author}{\bibinfo{person}{Kai Wang}, \bibinfo{person}{Xiaojiang Peng}, \bibinfo{person}{Jianfei Yang}, \bibinfo{person}{Shijian Lu}, {and} \bibinfo{person}{Yu Qiao}.} \bibinfo{year}{2020}\natexlab{b}.
\newblock \showarticletitle{Suppressing uncertainties for large-scale facial expression recognition}. In \bibinfo{booktitle}{\emph{Proceedings of the IEEE/CVF conference on computer vision and pattern recognition}}. \bibinfo{pages}{6897--6906}.
\newblock


\bibitem[Xu et~al\mbox{.}(2016)]%
        {13_xu2016video}
\bibfield{author}{\bibinfo{person}{Baohan Xu}, \bibinfo{person}{Yanwei Fu}, \bibinfo{person}{Yu-Gang Jiang}, \bibinfo{person}{Boyang Li}, {and} \bibinfo{person}{Leonid Sigal}.} \bibinfo{year}{2016}\natexlab{}.
\newblock \showarticletitle{Video emotion recognition with transferred deep feature encodings}. In \bibinfo{booktitle}{\emph{proceedings of the 2016 ACM on international conference on multimedia retrieval}}. \bibinfo{pages}{15--22}.
\newblock


\bibitem[Xu et~al\mbox{.}(2023)]%
        {53_xu2023mmcosine}
\bibfield{author}{\bibinfo{person}{Ruize Xu}, \bibinfo{person}{Ruoxuan Feng}, \bibinfo{person}{Shi-Xiong Zhang}, {and} \bibinfo{person}{Di Hu}.} \bibinfo{year}{2023}\natexlab{}.
\newblock \showarticletitle{MMCosine: Multi-Modal Cosine Loss Towards Balanced Audio-Visual Fine-Grained Learning}. In \bibinfo{booktitle}{\emph{ICASSP 2023-2023 IEEE International Conference on Acoustics, Speech and Signal Processing (ICASSP)}}. IEEE, \bibinfo{pages}{1--5}.
\newblock


\bibitem[Xue et~al\mbox{.}(2022)]%
        {new_12_xue2022coarse}
\bibfield{author}{\bibinfo{person}{Fanglei Xue}, \bibinfo{person}{Zichang Tan}, \bibinfo{person}{Yu Zhu}, \bibinfo{person}{Zhongsong Ma}, {and} \bibinfo{person}{Guodong Guo}.} \bibinfo{year}{2022}\natexlab{}.
\newblock \showarticletitle{Coarse-to-fine cascaded networks with smooth predicting for video facial expression recognition}. In \bibinfo{booktitle}{\emph{Proceedings of the IEEE/CVF Conference on Computer Vision and Pattern Recognition}}. \bibinfo{pages}{2412--2418}.
\newblock


\bibitem[Yang et~al\mbox{.}(2024)]%
        {new_55_yang2024emogen}
\bibfield{author}{\bibinfo{person}{Jingyuan Yang}, \bibinfo{person}{Jiawei Feng}, {and} \bibinfo{person}{Hui Huang}.} \bibinfo{year}{2024}\natexlab{}.
\newblock \showarticletitle{EmoGen: Emotional Image Content Generation with Text-to-Image Diffusion Models}. In \bibinfo{booktitle}{\emph{Proceedings of the IEEE/CVF Conference on Computer Vision and Pattern Recognition}}. \bibinfo{pages}{6358--6368}.
\newblock


\bibitem[Yang et~al\mbox{.}(2021)]%
        {new_3_yang2021circular}
\bibfield{author}{\bibinfo{person}{Jingyuan Yang}, \bibinfo{person}{Jie Li}, \bibinfo{person}{Leida Li}, \bibinfo{person}{Xiumei Wang}, {and} \bibinfo{person}{Xinbo Gao}.} \bibinfo{year}{2021}\natexlab{}.
\newblock \showarticletitle{A circular-structured representation for visual emotion distribution learning}. In \bibinfo{booktitle}{\emph{Proceedings of the IEEE/CVF Conference on Computer Vision and Pattern Recognition}}. \bibinfo{pages}{4237--4246}.
\newblock


\bibitem[You et~al\mbox{.}(2016)]%
        {new_22_you2016building}
\bibfield{author}{\bibinfo{person}{Quanzeng You}, \bibinfo{person}{Jiebo Luo}, \bibinfo{person}{Hailin Jin}, {and} \bibinfo{person}{Jianchao Yang}.} \bibinfo{year}{2016}\natexlab{}.
\newblock \showarticletitle{Building a large scale dataset for image emotion recognition: The fine print and the benchmark}. In \bibinfo{booktitle}{\emph{Proceedings of the AAAI conference on artificial intelligence}}, Vol.~\bibinfo{volume}{30}.
\newblock


\bibitem[Yu et~al\mbox{.}(2022)]%
        {new_37_yu2022mm}
\bibfield{author}{\bibinfo{person}{Jiashuo Yu}, \bibinfo{person}{Ying Cheng}, \bibinfo{person}{Rui-Wei Zhao}, \bibinfo{person}{Rui Feng}, {and} \bibinfo{person}{Yuejie Zhang}.} \bibinfo{year}{2022}\natexlab{}.
\newblock \showarticletitle{Mm-pyramid: Multimodal pyramid attentional network for audio-visual event localization and video parsing}. In \bibinfo{booktitle}{\emph{Proceedings of the 30th ACM international conference on multimedia}}. \bibinfo{pages}{6241--6249}.
\newblock


\bibitem[Zadeh et~al\mbox{.}(2018)]%
        {16_zadeh2018multimodal}
\bibfield{author}{\bibinfo{person}{AmirAli~Bagher Zadeh}, \bibinfo{person}{Paul~Pu Liang}, \bibinfo{person}{Soujanya Poria}, \bibinfo{person}{Erik Cambria}, {and} \bibinfo{person}{Louis-Philippe Morency}.} \bibinfo{year}{2018}\natexlab{}.
\newblock \showarticletitle{Multimodal language analysis in the wild: Cmu-mosei dataset and interpretable dynamic fusion graph}. In \bibinfo{booktitle}{\emph{Proceedings of the 56th Annual Meeting of the Association for Computational Linguistics}}. \bibinfo{pages}{2236--2246}.
\newblock


\bibitem[Zhang et~al\mbox{.}(2023)]%
        {40_zhang2023weakly}
\bibfield{author}{\bibinfo{person}{Zhicheng Zhang}, \bibinfo{person}{Lijuan Wang}, {and} \bibinfo{person}{Jufeng Yang}.} \bibinfo{year}{2023}\natexlab{}.
\newblock \showarticletitle{Weakly Supervised Video Emotion Detection and Prediction via Cross-Modal Temporal Erasing Network}. In \bibinfo{booktitle}{\emph{Proceedings of the IEEE/CVF Conference on Computer Vision and Pattern Recognition}}. \bibinfo{pages}{18888--18897}.
\newblock


\bibitem[Zhang et~al\mbox{.}(2024)]%
        {new_59_zhang2024mart}
\bibfield{author}{\bibinfo{person}{Zhicheng Zhang}, \bibinfo{person}{Pancheng Zhao}, \bibinfo{person}{Eunil Park}, {and} \bibinfo{person}{Jufeng Yang}.} \bibinfo{year}{2024}\natexlab{}.
\newblock \showarticletitle{Mart: Masked affective representation learning via masked temporal distribution distillation}. In \bibinfo{booktitle}{\emph{Proceedings of the IEEE/CVF Conference on Computer Vision and Pattern Recognition}}. \bibinfo{pages}{12830--12840}.
\newblock


\bibitem[Zhao et~al\mbox{.}(2020)]%
        {35_zhao2020end}
\bibfield{author}{\bibinfo{person}{Sicheng Zhao}, \bibinfo{person}{Yunsheng Ma}, \bibinfo{person}{Yang Gu}, \bibinfo{person}{Jufeng Yang}, \bibinfo{person}{Tengfei Xing}, \bibinfo{person}{Pengfei Xu}, \bibinfo{person}{Runbo Hu}, \bibinfo{person}{Hua Chai}, {and} \bibinfo{person}{Kurt Keutzer}.} \bibinfo{year}{2020}\natexlab{}.
\newblock \showarticletitle{An end-to-end visual-audio attention network for emotion recognition in user-generated videos}. In \bibinfo{booktitle}{\emph{Proceedings of the AAAI Conference on Artificial Intelligence}}, Vol.~\bibinfo{volume}{34}. \bibinfo{pages}{303--311}.
\newblock


\bibitem[Zhao and Liu(2021)]%
        {new_34_zhao2021former}
\bibfield{author}{\bibinfo{person}{Zengqun Zhao} {and} \bibinfo{person}{Qingshan Liu}.} \bibinfo{year}{2021}\natexlab{}.
\newblock \showarticletitle{Former-dfer: Dynamic facial expression recognition transformer}. In \bibinfo{booktitle}{\emph{Proceedings of the 29th ACM International Conference on Multimedia}}. \bibinfo{pages}{1553--1561}.
\newblock


\bibitem[Zhao and Patras(2023)]%
        {new_36_zhao2023prompting}
\bibfield{author}{\bibinfo{person}{Zengqun Zhao} {and} \bibinfo{person}{Ioannis Patras}.} \bibinfo{year}{2023}\natexlab{}.
\newblock \showarticletitle{Prompting visual-language models for dynamic facial expression recognition}.
\newblock \bibinfo{journal}{\emph{arXiv preprint arXiv:2308.13382}} (\bibinfo{year}{2023}).
\newblock


\bibitem[Zhou et~al\mbox{.}(2019)]%
        {new_4_zhou2019exploring}
\bibfield{author}{\bibinfo{person}{Hengshun Zhou}, \bibinfo{person}{Debin Meng}, \bibinfo{person}{Yuanyuan Zhang}, \bibinfo{person}{Xiaojiang Peng}, \bibinfo{person}{Jun Du}, \bibinfo{person}{Kai Wang}, {and} \bibinfo{person}{Yu Qiao}.} \bibinfo{year}{2019}\natexlab{}.
\newblock \showarticletitle{Exploring emotion features and fusion strategies for audio-video emotion recognition}. In \bibinfo{booktitle}{\emph{2019 International conference on multimodal interaction}}. \bibinfo{pages}{562--566}.
\newblock


\end{thebibliography}

\end{document}